\documentclass{article}  

\usepackage{arxiv}

\usepackage[utf8]{inputenc} 
\usepackage[T1]{fontenc}    
\usepackage{hyperref}       
\usepackage{url}            
\usepackage{booktabs}       
\usepackage{amsfonts}       
\usepackage{nicefrac}       
\usepackage{microtype}      
\usepackage{graphicx}
\usepackage{natbib}
\usepackage{doi}
\usepackage{longtable}
\usepackage{array}
\usepackage{multirow}
\usepackage{amsmath}
\usepackage{cleveref}       
\usepackage{algorithm}
\usepackage{algpseudocode}
\usepackage{amsmath}
\usepackage{geometry}
\usepackage{float}  
\usepackage{appendix}
\usepackage[font=normalsize]{caption}
\usepackage{ragged2e}
\usepackage{fancyhdr}
\title{ScaleMCP: Dynamic and Auto-Synchronizing Model Context Protocol Tools for LLM Agents}

\newif\ifuniqueAffiliation
\uniqueAffiliationtrue

\ifuniqueAffiliation
\author{
    \textbf{Elias Lumer}\thanks{elias.lumer@pwc.com \quad * anmol.b.gulati@pwc.com}, 
    \textbf{Anmol Gulati}\footnotemark[1], 
    \textbf{Vamse Kumar Subbiah}, \\
    \textbf{Pradeep Honaganahalli Basavaraju}, 
    \textbf{James A. Burke} \\
    \textit{PricewaterhouseCoopers, U.S.A.} \\
}

\else
\fi

\hypersetup{
pdftitle={ScaleMCP: Dynamic, Scalable, and Auto-Synchronizing Model Context Protocol Tools for LLM Agents},
pdfsubject={cs.AI},
pdfauthor={Elias Lumer, Anmol Gulati, Vamse Kumar Subbiah, Pradeep Honaganahalli Basavaraju, James A. Burke},
pdfkeywords={Retrieval-Augmented Generation, Model Context Protocol, Tool Retrieval, LLMs, Agents},
}

\begin{document}
\maketitle
\begin{abstract}
Recent advancements in Large Language Models (LLMs) and the introduction of the Model Context Protocol (MCP) have significantly expanded LLM agents' capability to interact dynamically with external tools and APIs. However, existing tool selection frameworks do not integrate MCP servers, instead relying heavily on error-prone manual updates to monolithic local tool repositories, leading to duplication, inconsistencies, and inefficiencies. Additionally, current approaches abstract tool selection before the LLM agent is invoked, limiting its autonomy and hindering dynamic re-querying capabilities during multi-turn interactions. To address these issues, we introduce ScaleMCP, a novel tool selection approach that dynamically equips LLM agents with a MCP tool retriever, giving agents the autonomy to add tools into their memory, as well as an auto-synchronizing tool storage system pipeline through CRUD (create, read, update, delete) operations with MCP servers as the single source of truth. We also propose a novel embedding strategy, Tool Document Weighted Average (TDWA), designed to selectively emphasize critical components of tool documents (e.g. tool name or synthetic questions) during the embedding process. Comprehensive evaluations conducted on a created dataset of 5,000 financial metric MCP servers, across 10 LLM models, 5 embedding models, and 5 retriever types, demonstrate substantial improvements in tool retrieval and agent invocation performance, emphasizing ScaleMCP’s effectiveness in scalable, dynamic tool selection and invocation.
\end{abstract}


\section{Introduction}
Recent advancements in Large Language Models (LLMs) and tool learning have enabled LLM agents to dynamically interact with external tools and APIs. The introduction of the Model Context Protocol (MCP) standardizes this connection between LLMs and external tools, data sources, and prompts \citep{mcp_anthropic2024}. Concurrently, to deal with LLM architecture limitations in calling the correct tools or model providers not allowing more than 128 tools to be equipped to the LLM, breakthroughs in tool-applied Retrieval-Augmented Generation (RAG) have enabled LLM agents to efficiently scale to a large number of tools  \citep{lumericaart25,chen2024reinvoketoolinvocationrewriting}.

\begin{figure}[htbp]
  \centering
  \includegraphics[width=1\linewidth]{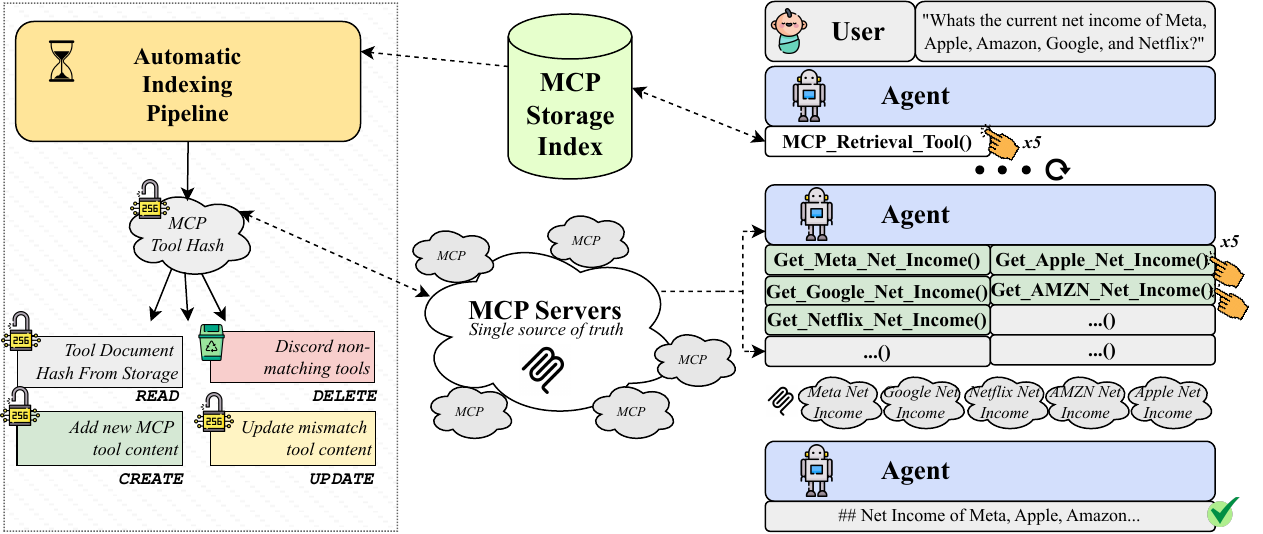}
  \caption{ScaleMCP automatic indexing pipeline and LLM agent invocation. The auto-synchronization tool indexing pipeline reads the current single source of truth MCP server tools and compares its hashes to the MCP storage system hashes, with CRUD (create, read, update, delete) operations on the storage index. For the LLM agent invocation, after the user asks a question, the LLM agent calls the "MCP Retrieval Tool" in parallel 5 times (1 for each targeted tool to retrieve) to equip relevant MCP servers (tools) into its context. Then, after retrieving the relevant MCPs, the LLM agent decides to call 5 MCP servers in parallel, and the MCP server returns the tool response. Finally, the LLM agent returns a successful final answer to the user after reasoning through the MCP server responses.}\label{fig:scalemcp}
\end{figure}

Despite technological advancements in tool selection and LLM invocation in prior work, three critical limitations remain. First, existing frameworks have not adopted the Model Context Protocol within their tool selection frameworks. Second, prior work heavily relies on manual updates to a monolithic tool repository to maintain synchronization between tool definitions and tool storage systems used for retrieval. This manual updating process is prone to human-error, inconsistencies, and relies on duplicated tool code. Lastly, current approaches abstract the tool selection process outside of the LLM invocation process, limiting the agents' autonomy and preventing dynamic re-querying of the tool storage system during multi-turn user conversations.

In this paper, we address these gaps by introducing ScaleMCP, a novel tool selection approach enabling LLM agents to dynamically discover and equip MCPs (as tools) during multi-turn interactions. Central to our framework is an auto-synchronization tool storage system pipeline (Figure \ref{fig:scalemcp}) that treats MCP servers as the single source of truth, automatically detecting and reflecting updates in the storage system by using CRUD operations (create, read, update, delete). Additionally, we propose a new tool document embedding process, Tool Document Weighted Average (TDWA), which selectively emphasizes certain tool document components during vector embedding, overcoming the current concatenation or simple averaging method that treats all components equally in the vector space. Lastly, we validate our contributions with a newly curated dataset of 5,000 real-world financial metric MCP servers, with extensive experiments on 10 LLM models, 5 embedding models, and 5 retriever types for (1) tool retrieval, (2) comparison of TDWA, and (3) end-to-end LLM agent invocation. 

\section{Background}

\subsection{Model Context Protocol (MCP)}
The Model Context Protocol (MCP) is an open protocol developed to standardize the integration between Large Language Models (LLMs) and external tools, data sources, and prompts \citep{mcp_anthropic2024}. Anthropic introduced MCP to provide a universal method to replace fragmented integrations with a unified protocol for AI agents. This architecture allows developers to expose their tools, APIs, data, or prompts through MCP servers or build AI applications (MCP clients) that connect to these servers, simplifying the process of granting AI systems access to necessary data. Recently, researchers have highlighted security and privacy considerations associated with the Model Context Protocol, such as malicious code execution, remote access control, credential theft, lack of authentication, authorization, and debugging \citep{radosevich2025mcpsafetyauditllms,hou2025modelcontextprotocolmcp}. Additionally, developers have noted a limitation with serverless deployments as MCP currently is a stateful protocol between a client and server, with benefits including push notifications and sampling \citep{volobuilds_mcp2025}. Nonetheless, the Model Context Protocol is asserting itself as the current standard for LLM agent tool integration, with model providers (e.g. OpenAI and Google) and AI platforms (e.g. Cursor, Cline) adopting the protocol \citep{openai_2025,google_gemini_2025,cursor_2025,cline_2025}. 

\subsection{Tool Selection and Retrieval}
Large language models natively face limitations in the number of tools or functions they can directly access and invoke. On one hand, complex multi-hop tool usage constrains the reasoning capability of the LLM in deciding which tools to invoke and in what sequence. On the other hand, model providers such as OpenAI, Anthropic, and Google enforce strict API limits, preventing the integration of more than 128 tools at a time \citep{openaifunctioncalling2024}. To scale beyond this constraint, prior works \citep{lumericaart25,chen2024reinvoketoolinvocationrewriting} employ advanced RAG-based methods without fine-tuning, storing tools offline in vector databases or knowledge graphs \citep{lumer2025graphragtoolfusion,peng2024graphretrievalaugmentedgenerationsurvey} and dynamically equipping only relevant tools during inference. Alternatively, an agentic RAG approach equips LLMs with dedicated tool-searching functionalities, allowing dynamic self-directed tool discovery and invocation \citep{singh2025agenticretrievalaugmentedgenerationsurvey, li2023apibankcomprehensivebenchmarktoolaugmented, du2024anytoolselfreflectivehierarchicalagents}, contrasting with static, predefined retrieval pipelines \citep{lumer2024toolshedscaletoolequippedagents,chen2024reinvoketoolinvocationrewriting}. However, \citeauthor{li2023apibankcomprehensivebenchmarktoolaugmented} notes a limitation where earlier GPT-based models fail to utilize these dynamic tool-search functions effectively. Additional research emphasizes retriever fine-tuning over out-of-the-box embeddings provided by OpenAI \citep{openai_2025} or Google \citep{google_gemini_2025} for tool selection efficiency \citep{wu2024sealtoolsselfinstructtoollearning,qin2023toolllmfacilitatinglargelanguage,anantha2023protipprogressivetoolretrieval,yuan2024craftcustomizingllmscreating,zheng2024toolrerankadaptivehierarchyawarereranking}. Underlying tool retrieval methods vary from lexical-based keyword searches \citep{robertson_probabilistic_2009} to vector-based and graph-based strategies \citep{gao2024retrievalaugmentedgenerationlargelanguage,peng2024graphretrievalaugmentedgenerationsurvey}, reflecting diverse advancements within the RAG paradigm. ScaleMCP uses a hybrid approach combining prior work \citep{lumericaart25,chen2024reinvoketoolinvocationrewriting,li2023apibankcomprehensivebenchmarktoolaugmented,du2024anytoolselfreflectivehierarchicalagents}, using out-of-the-box embeddings and LLMs with advanced RAG or Graph-RAG retrieval strategies for tool storage, and equipping an LLM agent with a MCP-retrieval tool. Unlike previous approaches that rely on simple embedding concatenation or averaging methods \citep{chen2024reinvoketoolinvocationrewriting,lumericaart25}, our Tool Document Weighted Average (TDWA) can dynamically weight individual components of tool documents, which prevents over-emphasis on certain tool document components. Furthermore, ScaleMCP solves the limitations associated with static, monolithic local tool repositories and non-automatic updates \citep{lumericaart25}, leveraging the underlying bidirectional Model Context Protocol connection as a medium for efficient tool execution and mapping post-retrieval.
\subsection{Tool Calling for LLM Inovcation}
While tool selection involves curating the relevant tools to equip the LLM agent, prior work also focuses on the pure LLM tool invocation \citep{hao2024toolkengptaugmentingfrozenlanguage,qin2023toolllmfacilitatinglargelanguage,patil2023gorillalargelanguagemodel}. Additionally, modern finetuning approaches for LLM tool calling include MOLoRA \citep{hao2024citienhancingtoolutilizing}, efficient tree-based methods \citep{zhu2025dividethenaggregateefficienttoollearning}, or curating high quality tool-instruction datasets using multi-AI agents \citep{liu2024toolacewinningpointsllm,zhuang2025hephaestusimprovingfundamentalagent}. While finetuning LLMs is a promising area of research for LLM tool learning, the focus of this paper is a plug-and-play method using out-of-the-box LLMs and embeddings from OpenAI, Google, Anthropic, and Meta \citep{openai_2025,google_gemini_2025,anthropic_2025,meta_llama_2025}. Furthermore, as stated previously, we use an agentic-RAG-inspired approach that equips an LLM with an MCP retriever tool, giving the LLM the autonomy in the LLM tool selection and invocation.

\section{Methodology}

\subsection{ScaleMCP Overview}
We introduce ScaleMCP, a novel approach to LLM agent tool selection for MCP servers (tools), encompassing an auto-synchronizing tool storage system indexing pipeline and a modern agentic RAG approach that gives the tool invocation autonomy to the LLM agent (See Figure \ref{fig:scalemcp}). By using the built-in function-calling capability of the LLM, ScaleMCP allows LLM agents to have access and use thousands of MCP servers, autonomously managing its tool storage which the underlying tool storage system is synced to the available MCP servers automatically. 

\begin{algorithm}
\caption{ScaleMCP Auto-Synchronization Indexing Pipeline}\label{alg:scalemcp}
\begin{algorithmic}[1]
\Require MCP Server Tool List $M$, Existing Storage System Hashes $S$
\State Initialize empty sets: $to\_index$, $seen\_hashes$
\ForAll{tool $m \in M$}
    \State $content \gets m.tool\_name \| m.tool\_description \| m.tool\_arguments$
    \State $hash \gets \text{SHA256}(content)$
    \State $seen\_hashes.add(hash)$
    \If{$hash \notin S$}
        \State $to\_index.add(m)$
    \EndIf
\EndFor

\State \textbf{Remove outdated tools:}
\ForAll{stored\_tool $s \in S$}
    \If{$s \notin seen\_hashes$}
        \State Remove storage entry corresponding to hash $s$
    \EndIf
\EndFor

\State \textbf{Index new or updated tools:}
\ForAll{tool $t \in to\_index$}
    \State Perform Storage-specific Mapping Function (generalized)
    \State Store new tool hash and indexing outputs into the storage system
\EndFor
\end{algorithmic}
\end{algorithm}

\subsection{ScaleMCP Auto-Synchronization Indexing Pipeline}
The tool storage system can be chosen by the tool selection use case and the retrieval method. While the most common storage system is a vector database and vector retrieval, other options include graph database, a hybrid graph RAG approach, or lexical term matching in a standard database. For example, if MCP servers are independent of each other, a vector database can scale independently. If MCP servers are dependent on each other in a graphical manner, a graph database can scale this dependency information for retrieval. ScaleMCP is driven by its auto-synchronization tool storage indexing pipeline, which uses the MCP servers as the single source of truth to determine any net new CRUD (create, read, update, delete) operations to the tool storage system. In Algorithm \ref{alg:scalemcp}, first the total MCP tools are retrieved and a SHA-256 hash is computed using the tool name, description, and parameters. The new MCP tool hashes are compared to the existing hashes in the existing storage system. If both hashes match, nothing occurs. If a hash does not match, the existing storage system tool is discarded, and the new MCP tool is added using the storage-specific mapping function (and hashed). This storage-specific mapping function can be the embedding function (optionally TDWA, see Figure \ref{fig:tdwa}) for a vector database, the nodes and edges calculation of a graph database, or a simple lexical index. 

\subsubsection{New Embedding Mapping Function Tool Document Weighted Average}
\begin{figure}[htbp]
  \centering
  \includegraphics[width=0.7\linewidth]{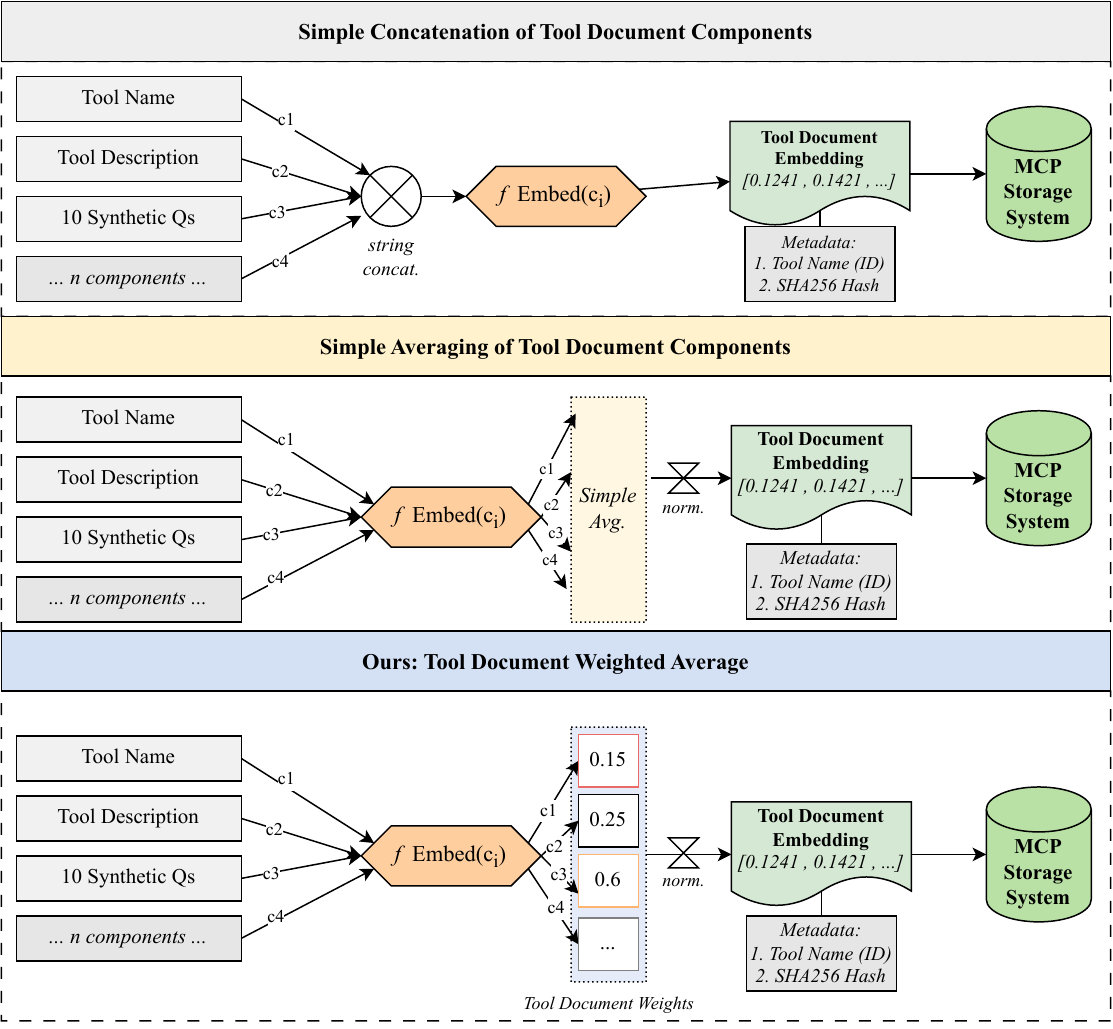}
  \caption{Tool‐Document Weighted Average Embedding comparison to simple concatenation\label{fig:tdwa}}
\end{figure}

Prior work \citep{lumer2025graphragtoolfusion,chen2024reinvoketoolinvocationrewriting} uses only a concatenation or simple average of tool document components, or simple tool descriptions \citep{yuan2024easytoolenhancingllmbasedagents,anantha2023protipprogressivetoolretrieval}. Tool document components can be classified as features of a tool -- tool name, description, parameters -- or augmentations such as synthetic questions or key topics \citep{gao2024retrievalaugmentedgenerationlargelanguage} about the specific tool. We introduce a new embedding function paradigm for tool-specific use cases, Tool Document Weighted Average, which allows the importance of each tool document component to contribute to a weighted average of embeddings. In Figure \ref{fig:tdwa}, we compare our Tool Document Weighted Average embedding approach to simple concatenation or averaging of tool document components.
\begin{equation}
\label{eq:td-weighted-average}
\mathbf{z}_{\mathrm{ToolDocument_WA}}
=
\frac{\displaystyle\sum_{i=1}^N w_i\,\mathrm{Embed}\bigl(c_i\bigr)}
     {\displaystyle\Bigl\|\sum_{i=1}^N w_i\,\mathrm{Embed}\bigl(c_i\bigr)\Bigr\|_2}
\end{equation}
In Equation~\ref{eq:td-weighted-average}, we decompose a tool document into $N$ tool document components $c_i$ (e.g. tool name, description, $S$ synthetic questions), each assigned a nonnegative weight $w_i$ with $\sum_{i=1}^N w_i = 1$. We then compute the weighted sum of their embeddings $\mathrm{Embed}(c_i)$ and normalize the vector to unit length. This normalized, weighted‐average tool document embedding both preserves the relative importance assigned to each component, providing fine‐grained control beyond simple concatenation or unweighted averaging. 

\subsection{ScaleMCP LLM Invocation}
To enable scalable tool selection within the LLM invocation, we equip the LLM agent with a specialized MCP Retrieval tool, which the LLM can pass in keywords to retrieve relevant MCP servers. As seen in Figure \ref{fig:scalemcp}, when an LLM agent uses the MCP Retrieval Tool (in the example, the MCP Retrieval Tool is called 5 times with 5 sets of keywords related to the net income to each financial company), the framework automatically loads the retrieved MCP servers into the context of the LLM and "binds" the new tools to the LLM \citep{openaifunctioncalling2024} using function calling. Then, when the LLM agent sees the new MCP servers it has access to, it can call the tools in parallel to get responses from the MCP servers. Finally, the LLM reasons from the MCP server tool responses a final answer to the user. The benefit to giving autonomy to the LLM agent to use a MCP Retrieval tool is that it can continue to re-query the MCP storage system if no successful matches are found. Furthermore, the LLM is able to better manage its own tool memory for multi-turn chats where it knows what tools it had access to in the session, and when to query for new MCP servers. The benefit of MCP servers in the LLM invocation case is the standardization of tool calling and the ecosystem of MCP servers to connect to.
\section{Dataset Construction}
To evaluate the capabilities of ScaleMCP, we created a large-scale real-world dataset consisting of 5,000 company-based financial metric MCP servers and a corresponding set of instances or user queries with expected tool calls. Our dataset was built to simulate realistic agent-tool interactions over financial metrics, while remaining cost-efficient and reproducible.

\subsection{Tool Creation}
We began with the Fortune 1000 companies and generated five deterministic tools for each company \citep{investopediafortune10002025}.
\begin{itemize}
    \item \texttt{get\_\{company\}\_current\_stock\_price}
    \item \texttt{get\_\{company\}\_stock\_price\_history}
    \item \texttt{get\_\{company\}\_analyst\_price\_targets}
    \item \texttt{get\_\{company\}\_revenue}
    \item \texttt{get\_\{company\}\_net\_income}
\end{itemize}

These tools were implemented using the open-source yfinance Python package \citep{aroussi2025yfinance}. This API was used solely for academic research purposes; the tools are not designed for production or commercial use. All tool definitions were generated programmatically using deterministic templates. Tool names, descriptions, and parameter schemas were auto-filled using structured metadata such as company name, ticker symbol, and fiscal year. No LLMs were involved in the creation of the tools themselves. Each tool was served via an MCP-compliant server using the open-source \texttt{fast-mcp} framework \citep{jlowinfastmcp2025}, resulting in exactly 5,000 MCP servers.

\begin{table*}[t]
\centering
\small  
\caption{ScaleMCP Dataset and Evaluation Statistics.}
\label{tab:dataset_stats}
\begin{tabular}{@{}lc@{}}
\toprule
\textbf{Metric} & \textbf{Value} \\
\midrule
Number of MCP Servers & 5,000 \\
Number of User Query Instances & 140,000 \\
Average Tool Calls per Instance & 5.0 \\
\bottomrule
\end{tabular}
\end{table*}

\subsubsection{Tool Document Synthetic Question Creation}
To improve representation of each tool in vector space, we enriched the tool documents with synthetic natural language questions. For each of the five tool templates, we generated either 0, 5, or 10 synthetic questions using an LLM. Each question was created by conditioning on a generic template with a \texttt{\{company\}} placeholder. We then filled this placeholder with either the company name, its stock ticker, or an alternate company alias \citep{investopediafortune10002025}. This approach introduced variation in phrasing and surface form while remaining semantically faithful to the tool’s functionality. The resulting documents simulate realistic queries agents might encounter, making them better suited for dense retrieval and reranking tasks.

\subsection{User Query Instance Generation}
In addition to synthetic questions embedded in tool documents, we created a set of standalone user queries designed to evaluate retrieval performance and agent reasoning. These queries were modeled after the tool templates but crafted to simulate natural user prompts, often involving implicit reasoning or multi-hop dependencies. Rather than generating queries for each company individually, we created approximately 100 base queries per tool and then templated them across all 1,000 companies. This reduced LLM inference cost while still producing a large and diverse evaluation set. The final dataset includes approximately 140,000 user query instances, covering a wide range of financial tasks across companies, tools, and phrasings.

\section{Evaluations}

\subsection{Experiment 1: MCP Vector Database Retrieval}
\subsubsection{Experiment Settings}
We evaluate the retrieval effectiveness of different embedding models in retrieving relevant MCP tool documents. Our evaluation is conducted on a dataset of 5,000 MCP servers using a simple concatenation strategy to store tool representations. In total, we test five embedding models: \texttt{OpenAI text-embedding-3-large}, \texttt{OpenAI text-embedding-3-small}, \texttt{OpenAI text-embedding-ada-002}, \texttt{Amazon titan-embed-text-v1}, and \texttt{VertexAI text-embedding-005}. Each model is evaluated under six search configurations: vector-only search, hybrid search, BM25 lexical search, reranking using Cohere’s cross-encoder reranker (v3-english), and LLM-based rerankers (GPT-4o and Claude Sonnet 3.7). We also vary the number of synthetic questions embedded in the tool documents (SQ = 0, 5, or 10) and report NDCG, Recall, and MAP at $K=1$, $K=5$, and $K=10$. For brevity, we highlight only the $K=5$ results and a subset of three embedding models below; full results across all embedding models and $K$ values are provided in Appendix Table~\ref{tab:retrieval_performance_2}.

\subsubsection{Results Analysis}
Table~\ref{tab:retrieval_performance} summarizes retrieval performance across all embedding models and search configurations. Vector-only retrieval performs poorly across all models, with MAP values in the 0.50 range, even for top-performing embeddings. This is attributed to the nature of the evaluation queries, which are predominantly multi-hop — requiring retrieval of multiple golden tools per question. Prior work has shown that dense retrieval struggles under multi-hop supervision, as vector similarity cannot simultaneously capture multiple tool intents with a single embedding \citep{wu2024sealtoolsselfinstructtoollearning}.

Despite this, reranking strategies show clear improvements. Cohere's cross-encoder reranker (v3-english) significantly boosts performance over vector-only baselines, while GPT-4o and Claude Sonnet 3.7 rerankers yield the strongest scores overall. The best Recall@10 (0.94) is achieved using \texttt{VertexAI text-embedding-005} with GPT-4o reranking, and the best MAP@10 (0.59) is obtained using the same embedding with Claude. BM25 remains the weakest performer across all metrics. Synthetic question enrichment (SQ=10) consistently improves retrieval results across configurations.


\begingroup
\scriptsize  

\begin{longtable}{@{}p{3.5cm}>{\centering\arraybackslash}p{2.5cm}>{\centering\arraybackslash}p{2.5cm}>{\centering\arraybackslash}p{2.5cm}@{}}
\caption{Retrieval performance by $K$ and metric using the concatenation storage strategy. Metrics are shown as synthetic queries (SQ) 0/5/10. Bolded values are the highest overall; underlined are second highest.}%
\label{tab:retrieval_performance}\\
\toprule
\textbf{Embedding Model} & \textbf{NDCG} & \textbf{Recall} & \textbf{MAP} \\
\midrule
\endfirsthead

\multicolumn{4}{c}{\tablename\ \thetable{} -- continued from previous page} \\
\toprule
\textbf{Embedding Model} & \textbf{NDCG} & \textbf{Recall} & \textbf{MAP} \\
\midrule
\endhead

\midrule
\multicolumn{4}{r}{\textit{Continued on next page}}\\
\midrule
\endfoot

\bottomrule
\endlastfoot

\addlinespace
\multicolumn{4}{@{}l}{\textit{Vector Search}}\\
\hspace*{1em}OpenAI text-embedding-3-large   & 0.62 / 0.59 / 0.63 & 0.83 / 0.85 / \underline{0.91} & 0.50 / 0.47 / 0.50 \\
\hspace*{1em}OpenAI text-embedding-3-small   & 0.64 / 0.65 / 0.64 & 0.82 / 0.84 / 0.85 & 0.53 / 0.54 / 0.53 \\
\hspace*{1em}VertexAI text-embedding-005       & 0.63 / 0.64 / 0.65 & 0.87 / 0.87 / \underline{0.91} & 0.51 / 0.51 / 0.51 \\

\addlinespace
\multicolumn{1}{@{}l}{\textit{Text Search (BM25)}} & 0.42 / 0.48 / 0.51 & 0.57 / 0.64 / 0.66 & 0.32 / 0.39 / 0.42 \\

\addlinespace
\multicolumn{4}{@{}l}{\textit{Hybrid Search}}\\
\hspace*{1em}OpenAI text-embedding-3-large   & 0.53 / 0.57 / 0.63 & 0.72 / 0.76 / 0.84 & 0.42 / 0.45 / 0.50 \\
\hspace*{1em}OpenAI text-embedding-3-small   & 0.55 / 0.62 / 0.64 & 0.74 / 0.78 / 0.83 & 0.45 / 0.50 / 0.53 \\
\hspace*{1em}VertexAI text-embedding-005       & 0.58 / 0.59 / 0.63 & 0.75 / 0.78 / 0.83 & 0.47 / 0.47 / 0.51 \\

\addlinespace
\multicolumn{4}{@{}l}{\textit{Cohere Reranking (v3-english)}}\\
\hspace*{1em}OpenAI text-embedding-3-large   & 0.64 / 0.61 / 0.64 & 0.85 / 0.85 / 0.90 & 0.52 / 0.49 / 0.51 \\
\hspace*{1em}OpenAI text-embedding-3-small   & 0.63 / 0.65 / 0.64 & 0.83 / 0.85 / 0.83 & 0.52 / 0.54 / 0.53 \\
\hspace*{1em}VertexAI text-embedding-005       & 0.65 / 0.64 / 0.67 & 0.87 / 0.88 / 0.90 & 0.53 / 0.52 / 0.54 \\

\addlinespace
\multicolumn{4}{@{}l}{\textit{LLM Reranking (GPT-4o)}}\\
\hspace*{1em}OpenAI text-embedding-3-large   & 0.66 / 0.64 / 0.67 & 0.88 / 0.88 / 0.91 & 0.54 / 0.51 / 0.54 \\
\hspace*{1em}OpenAI text-embedding-3-small   & 0.64 / 0.65 / 0.67 & 0.84 / 0.85 / 0.87 & 0.53 / 0.54 / 0.56 \\
\hspace*{1em}VertexAI text-embedding-005       & 0.67 / 0.68 / \underline{0.70} & 0.87 / 0.90 / \textbf{0.94} & 0.56 / 0.55 / \underline{0.58} \\

\addlinespace
\multicolumn{4}{@{}l}{\textit{LLM Reranking (Claude Sonnet 3.7)}}\\
\hspace*{1em}OpenAI text-embedding-3-large   & 0.66 / 0.63 / 0.67 & 0.88 / 0.88 / 0.91 & 0.54 / 0.50 / 0.54 \\
\hspace*{1em}OpenAI text-embedding-3-small   & 0.66 / 0.66 / \underline{0.70} & 0.84 / 0.85 / 0.89 & 0.55 / 0.54 / \textbf{0.59} \\
\hspace*{1em}VertexAI text-embedding-005       & 0.67 / 0.69 / \textbf{0.71} & 0.87 / 0.89 / 0.93 & 0.55 / \underline{0.58} / 0.59 \\
\end{longtable}
\endgroup


\subsubsection{Discussion}
These results confirm a critical limitation of traditional vector search in multi-hop settings: a single query embedding often fails to capture multiple distinct retrieval targets. This is consistent with prior findings in Seal-Tools \citep{wu2024sealtoolsselfinstructtoollearning}, which demonstrate poor vector retrieval under multi-answer supervision. In our case, a single query references multiple (3-12) golden MCP tools  (e.g., revenue + net income), making it unlikely that a single vector representation will retrieve all relevant tools effectively.

This motivates our ScaleMCP framework, which enables agentic behavior via a MCP-searching tool equipped to the agent. By allowing the agent to decompose the query and retrieve tools iteratively, we can overcome vector retrieval limitations and handle multi-hop tool queries. Our findings suggest that LLM-based reranking is highly effective as a patch over vector recall, but is computationally expensive. In future work, we plan to evaluate whether dynamic agent-driven retrieval using ScaleMCP can match or exceed reranked performance — but at lower inference cost and higher transparency. Therefore, while embedding quality and document enrichment matter, architectural adaptations—such as agentic retrieval—may ultimately be necessary for high-recall, multi-hop tool invocation.

\subsection{Experiment 2: LLM Agent Evaluation}
\subsubsection{Experimental Settings}
We evaluate the end-to-end performance of 10 LLM agents across retrieval and tool invocation tasks using the DeepEval framework \citep{deepeval2025}. The models include OpenAI's \texttt{gpt-4.1}, \texttt{gpt-4o}, \texttt{gpt-4o-mini}, and \texttt{gpt-o4-mini}, as well as Anthropic's \texttt{Claude 3.7 Sonnet}. Each agent is tested under three retrieval configurations: (1) BM25 (text-only baseline), (2) vector search using TDWA embeddings, and (3) vector search with reranking using the Cohere reranker (\texttt{v3-english}).

All retrieval is performed at $k=5$, and the retrieved MCP tools are passed into the LLM using OpenAI-compatible function calling. The tool index used in all runs is based on the best-performing setup from Experiment 1: TDWA (var-2) with 10 synthetic questions per tool, embedded using \texttt{OpenAI text-embedding-3-large}.

\begin{equation}\label{eq:tcorr}
\text{Tool Correctness} =
\frac{\textit{Number of Correct Tool Calls}}{\textit{Total Tool Calls}}
\end{equation}

\begin{equation}\label{eq:tcomp}
\text{Task Completion Score} = \mathit{AlignmentScore}(\mathit{Task}, \mathit{Outcome})
\end{equation}

\paragraph{Metric Definitions.}
We report two key evaluation metrics: \textit{Tool Correctness} and \textit{Task Completion Score}. 

As shown in Equation~\ref{eq:tcorr}, Tool Correctness measures whether the agent invoked the correct tool, supplied valid input arguments, and correctly interpreted the tool's output. 

Task Completion Score (Equation~\ref{eq:tcomp}) evaluates whether the agent's final response successfully fulfills the user's original query, regardless of the specific tools used. This metric reflects end-to-end effectiveness and is computed via alignment scoring between the expected and generated outputs.

\subsubsection{Results}
Table~\ref{tab:agent_performance} summarizes the Tool Correctness and Task Completion performance of 6 LLM agents under three retrieval strategies: BM25 (Text Search), vector search, and vector search with Cohere reranking. For brevity, we highlight 6 representative LLMs below; the full results for all 10 models are provided in Appendix~\ref{tab:agent_performance_2}.

Among all models, \texttt{gpt-o3} achieved the highest Task Completion score of 94.4\% using vector search combined with Cohere reranking. Despite its lower Tool Correctness (36.1\%), \texttt{gpt-o3} demonstrated strong ability to produce plausible or complete answers. In contrast, \texttt{gpt-4o-mini} achieved the highest Tool Correctness at 54.0\% and a strong Task Completion score of 86.7\% under the same configuration, making it the most balanced performer overall.

Larger models like \texttt{gpt-4.1} and \texttt{gpt-4o} also performed reliably, while \texttt{Claude 3.7 Sonnet} underperformed—particularly in Tool Correctness, with only 23.1\% when reranking was used. Most agents exceeded 80\% Task Completion with reranking, but Tool Correctness remained modest (ranging from 23\% to 54\%), indicating that agents often produce acceptable outputs without precise tool use.

The use of a cross-encoder reranking model consistently improved both metrics across models, compared to vector-only or BM25 retrieval, highlighting the importance of semantic reranking in enhancing tool retrieval and downstream reasoning performance.

\begin{table}[t]
\scriptsize  
\renewcommand{\arraystretch}{1.15}

\begin{longtable}{%
    >{\raggedright\arraybackslash}p{2.3cm} 
    >{\raggedright\arraybackslash}p{3.3cm} 
    >{\centering\arraybackslash}p{2.8cm}   
    >{\centering\arraybackslash}p{2.8cm}}  
\caption{Agent-level Tool Correctness and Task Completion rates at $k=5$ using the concatenation strategy and 10 questions per tool. The “Vector Search + Cohere Reranker” configuration uses the Cohere Reranker (\texttt{v3-english}).}
\label{tab:agent_performance}\\
\toprule
\textbf{LLM Agent} & \textbf{Retrieval Method} & \textbf{Tool Correctness (\%)} & \textbf{Task Completion (\%)} \\
\midrule
\endfirsthead

\multicolumn{4}{c}{\tablename\ \thetable{} -- continued from previous page}\\
\toprule
\textbf{LLM Agent} & \textbf{Retrieval Method} & \textbf{Tool Correctness (\%)} & \textbf{Task Completion (\%)} \\
\midrule
\endhead

\midrule
\multicolumn{4}{r}{\textit{Continued on next page}}\\
\midrule
\endfoot

\bottomrule
\endlastfoot
GPT 4.1 & Text Search & 39.1 & 73.2 \\
GPT 4.1 & Vector Search & 45.0 & 82.4 \\
GPT 4.1 & Vector Search + Cohere Reranker & 44.8 & 85.8 \\
\midrule
GPT 4o & Text Search & 47.9 & 75.9 \\
GPT 4o & Vector Search & \underline{51.7} & 86.5 \\
GPT 4o & Vector Search + Cohere Reranker & \underline{51.7} & 83.9 \\
\midrule
GPT 4o-mini & Text Search & 49.9 & 81.2 \\
GPT 4o-mini & Vector Search & 50.7 & 86.5 \\
GPT 4o-mini & Vector Search + Cohere Reranker & \textbf{54.0} & 86.7 \\
\midrule
GPT o4-mini & Text Search & 37.5 & 84.1 \\
GPT o4-mini & Vector Search & 38.6 & 81.3 \\
GPT o4-mini & Vector Search + Cohere Reranker & 40.7 & 85.6 \\
\midrule
GPT o3 & Text Search & 36.1 & 78.9 \\
GPT o3 & Vector Search & 22.2 & \underline{88.9} \\
GPT o3 & Vector Search + Cohere Reranker & 36.1 & \textbf{94.4} \\
\midrule
Claude 3.7 Sonnet & Text Search & 23.9 & 42.9 \\
Claude 3.7 Sonnet & Vector Search & 28.5 & 73.2 \\
Claude 3.7 Sonnet & Vector Search + Cohere Reranker & 23.1 & 69.4 \\

\end{longtable}
\end{table}

\normalsize

\subsubsection{Discussion}
The results reveal a key limitation in current LLM-based tool reasoning: high-quality task outputs often mask low underlying Tool Correctness. The dataset used has very complex queries with at times 12 expected tools that need to be inferred by the LLM agent. Although \texttt{gpt-o3} achieved the best Task Completion score (94.4\%), its Tool Correctness remained low at 36.1\%. Conversely, \texttt{gpt-4o-mini} balanced both metrics well, with 54.0\% Tool Correctness and 86.7\% Task Completion. This discrepancy shows that LLMs can generate fluent, plausible outputs without consistently invoking the correct tools or providing accurate input parameters.

This issue is amplified in complex multi-hop queries, where agents must orchestrate multiple tool calls—sometimes over 10—in a single response. The current static retrieval paradigm, even with strong rerankers like Cohere, limits an agent’s ability to reason iteratively or revise plans mid-task. Agents often rely on a fixed top-$k$ context and single-shot tool invocation, which does not support correction or reflective reasoning. To address this, agent architectures must evolve beyond static tool retrieval. Our ScaleMCP framework introduces a retrieval-augmented planning loop, where agents can iteratively search, evaluate, and invoke tools over multiple steps. By integrating a tool-search tool within the agent, models gain the ability to dynamically fetch tools based on intermediate results and rethink strategies when gaps are identified. Future work will focus on incorporating reflective reasoning modules—such as Anthropic’s \textit{“think”} tool—into this retrieval loop \citep{anthropicthinktool2025}. Combining agentic search with deliberate reasoning could significantly improve both Tool Correctness and transparency in complex tool-based workflows, particularly for high-stakes domains requiring grounded multi-step decisions, or calling 12 tools in a single turn.


\subsection{Experiment 3: TDWA Weighting Evaluation}
\subsubsection{Experimental Settings}
In this experiment, we evaluate the impact of different tool document storage strategies on retrieval effectiveness. We compare three strategies using a fixed synthetic question count of $SQ=10$ per tool: (1) \textit{Concat}, a simple unweighted concatenation of all tool components; (2) \textit{TDWA var-1}, with weights [0.2, 0.2, 0.2, 0.4]; and (3) \textit{TDWA var-2}, with weights [0.2, 0.3, 0, 0.5]. These weights represent the proportion of influence given to each component of a tool document in the weighted embedding: 20\% to the tool name, 20–30\% to the description, 0–20\% to the parameter schema, and 40–50\% to the synthetic questions. Each variant is tested using dense vector search, BM25, and reranking pipelines including Cohere’s reranker (\texttt{v3-english}), GPT-4o, and Claude Sonnet 3.7. All configurations use \texttt{OpenAI text-embedding-3-large} to embed documents and evaluate retrieval at $K=1$, $K=5$, and $K=10$.

\subsubsection{Results}
Table~\ref{tab:weighting_strategies_k5} presents retrieval results across all storage strategies and reranking methods at $K=5$. While we evaluated performance at $K=1$, $K=5$, and $K=10$, we report only the $K=5$ results below for brevity; full results across all $K$ values are provided in Appendix Table~\ref{tab:weighting_strategies}. In plain vector search, the simple \textit{Concat} strategy outperforms both TDWA variants in NDCG and Recall, achieving the highest top-1 performance (0.634 NDCG, 0.912 Recall). However, when reranking is applied, the gap narrows substantially. TDWA var-2 performs competitively, particularly in MAP and Recall under Cohere and Claude rerankers. Notably, TDWA var-2 outperforms Concat in several reranked MAP@5 scores, suggesting improved relevance ordering in larger candidate pools. Across all settings, the retriever using LLM reranking (GPT-4o and Claude 3.7) consistently yield the highest absolute retrieval scores, with Claude + Concat achieving the top NDCG (0.672) and MAP (0.539), and GPT-4o + Concat achieving the best Recall (0.912).

\begin{table}[t]
\scriptsize
\begin{longtable}{@{}p{4.0cm}p{4.4cm}p{2.4cm}ccc@{}}
\caption{Retrieval Performance at $K=5$ Across Search Strategies ($SQ=10$). 
Bold indicates the best value, underline is second best. 
TDWA weight vectors: var-1 = [0.2, 0.2, 0.2, 0.4] (name/desc/params/SQ), 
var-2 = [0.2, 0.3, 0, 0.5]. In var-1, 20\% of the weight is assigned to each of tool name, description, and parameters, and 40\% to synthetic questions. 
All runs use 10 synthetic questions.}

\label{tab:weighting_strategies_k5}\\
\toprule
\textbf{Strategy} & \textbf{Embedding Model} & \textbf{Weights} & \textbf{NDCG} & \textbf{Recall} & \textbf{MAP} \\
\midrule
\endfirsthead

\multicolumn{6}{c}{\tablename\ \thetable{} -- continued from previous page} \\
\toprule
\textbf{Strategy} & \textbf{Embedding Model} & \textbf{Weights} & \textbf{NDCG} & \textbf{Recall} & \textbf{MAP} \\
\midrule
\endhead

\midrule
\multicolumn{6}{r}{\textit{Continued on next page}}\\
\midrule
\endfoot

\bottomrule
\endlastfoot

Vector Search & OpenAI text-embedding-3-large & Concat & 0.634 & \textbf{0.912} & 0.499 \\
Vector Search & OpenAI text-embedding-3-large & TDWA var-1 & 0.631 & 0.886 & 0.504 \\
Vector Search & OpenAI text-embedding-3-large & TDWA var-2 & 0.620 & 0.891 & 0.485 \\
\midrule

Text Search (BM25) & -- & BM25 & 0.492 & 0.674 & 0.396 \\
\midrule

Cohere Reranker & OpenAI text-embedding-3-large & Concat & 0.642 & 0.896 & 0.510 \\
Cohere Reranker & OpenAI text-embedding-3-large & TDWA var-1 & 0.644 & 0.855 & 0.528 \\
Cohere Reranker & OpenAI text-embedding-3-large & TDWA var-2 & 0.629 & 0.839 & 0.511 \\
\midrule

LLM Reranker (GPT-4o) & OpenAI text-embedding-3-large & Concat & \underline{0.669} & \underline{0.906} & \textbf{0.545} \\
LLM Reranker (GPT-4o) & OpenAI text-embedding-3-large & TDWA var-1 & 0.638 & 0.885 & 0.505 \\
LLM Reranker (GPT-4o) & OpenAI text-embedding-3-large & TDWA var-2 & 0.656 & 0.889 & 0.528 \\
\midrule

LLM Reranker (Claude 3.7) & OpenAI text-embedding-3-large & Concat & \textbf{0.672} & \textbf{0.912} & \underline{0.539} \\
LLM Reranker (Claude 3.7) & OpenAI text-embedding-3-large & TDWA var-1 & 0.638 & 0.885 & 0.508 \\
LLM Reranker (Claude 3.7) & OpenAI text-embedding-3-large & TDWA var-2 & 0.644 & 0.889 & 0.511 \\

\midrule
\end{longtable}
\end{table}

\subsubsection{Discussion}
Although Tool Document Weighted Average (TDWA) does not outperform simple concatenation in raw vector search, these results do not imply that TDWA is ineffective. The strength of the Concat strategy in our evaluation likely stems from the highly keyword-driven nature of our tool dataset (e.g., tool names with company tickers and financial metrics), which align closely with surface-level terms in user queries. Additionally, the synthetic queries embedded in tools and the user queries used for evaluation were generated using similar LLM prompting techniques, potentially biasing results in favor of Concat by over-aligning vector space representations. Future work can use a portion of human queries to not have synthetic query bias.

In contrast, TDWA allows for finer control over the semantic contribution of each tool component and may generalize better in real-world use where user queries deviate more significantly from synthetic training prompts. TDWA var-2, which reduces reliance on parameter fields and increases emphasis on description and synthetic questions, shows particularly strong reranked performance — suggesting that structure-aware weighting improves document relevance when paired with a scoring model. Furthermore, this suggests that synthetic queries drive accuracy more than other tool document components.

Overall, these findings highlight the importance of considering the retrieval setting when choosing a storage strategy. While Concat is highly effective in synthetic evaluations, TDWA may offer better interpretability and robustness in diverse deployment contexts. In future work, we plan to test TDWA with human-written queries and explore adaptive weighting schemes that dynamically adjust based on query characteristics.

\section{Conclusion}
Advancements in Large Language Models (LLMs) and the introduction of the Model Context Protocol (MCP) have significantly improved LLM agents' dynamic interaction with external tools. However, existing tool selection approaches continue to face challenges related to manual storage system synchronization, inefficiencies from local monolithic tool repositories, and limited agent autonomy. To overcome these issues, we introduce ScaleMCP, a framework that enables LLM agents to autonomously manage a vast amount of MCPs during multi-turn interactions. ScaleMCP leverages auto-synchronization to its storage system through CRUD (create, read, update, delete) operations-with MCP servers as the single source of truth. For indexing, we introduce a novel Tool Document Weighted Average (TDWA) embedding strategy that brings fine-grained control of tool document components in the vector space.  Additionally, we evaluate ScaleMCP on collection of 5,000 financial MCP servers, varying 10 LLM models, 5 embedding models, and 5 retriever types. Our contribution to the tool learning field pushes the boundary of LLM tool selection with MCP servers.

\section{Limitations and Future Work}
While ScaleMCP pushes the needle forward in a modern adaption of the tool selection field, several limitations remain. First, the Model Context Protocol (MCP) is still in its early stages of development. Its reliance on a stateful client-server architecture, although practical for certain applications, may pose scalability and flexibility challenges compared to stateless or serverless alternatives that better align with modern distributed system design. Future iterations of ScaleMCP could explore hybrid architectures or serverless integrations to mitigate these concerns. Second, current large language models (LLMs) were not explicitly trained to autonomously manage tool discovery, dynamic context handling, or cross-tool interaction. As a result, their effectiveness in orchestrating tool-based workflows through MCP may be inherently limited. Fine-tuning LLMs with targeted objectives—such as dynamic tool discovery, multi-tool reasoning, and context-sensitive retrieval—represents a promising direction for future research. Recent efforts like Anthropic’s “think” tool \citep{anthropicthinktool2025}, which equips LLM agents with a dedicated reflection phase before tool invocation, offer a compelling path forward. Incorporating such reflective reasoning modules into the retrieval loop could help agents better assess context, correct errors mid-process, and plan more effective multi-step workflows. Third, our evaluation dataset focused primarily on financial metrics, which, while substantial, limits the generalizability of the findings to other domains. Future studies should validate MCP-based retrieval and orchestration approaches across a broader set of domains, including healthcare, law, scientific research, and customer support. Finally, emerging standards such as Google’s Agent-to-Agent (A2A) protocol \citep{googlea2a2025} highlight the growing importance of inter-agent communication standards for multi-agent collaboration. Integrating ScaleMCP with protocols such as A2A, or designing MCP extensions that enable autonomous agent negotiation, discovery, and handoff, could significantly enhance its applicability in open, multi-agent ecosystems. We encourage future work to explore such integrations, enabling richer, more flexible agent collaboration patterns.

\section{Ethical Considerations}
This research was conducted in compliance with the ACM Code of Ethics. The dataset utilized in this paper was constructed using publicly available real-time financial data accessed via the Open Source Python library "yfinance." Our usage adheres to Yahoo's API terms, specifically for non-commercial academic research purposes. All data was handled according to Yahoo's stipulated storage limits, and appropriate attribution has been provided. Potential ethical risks include inadvertent misrepresentation or inaccuracies in financial data, highlighting the importance of domain expert validation for critical applications. Furthermore, no additional personnel or external labor was employed for dataset creation or testing, maintaining transparency and ethical research practices.

\bibliographystyle{unsrtnat}
\bibliography{references.bib}  

\begin{thebibliography}{39}
\providecommand{\natexlab}[1]{#1}
\providecommand{\url}[1]{\texttt{#1}}
\expandafter\ifx\csname urlstyle\endcsname\relax
  \providecommand{\doi}[1]{doi: #1}\else
  \providecommand{\doi}{doi: \begingroup \urlstyle{rm}\Url}\fi

\bibitem[Anthropic(2024)]{mcp_anthropic2024}
Anthropic.
\newblock Introducing the model context protocol, 2024.
\newblock URL \url{https://www.anthropic.com/news/model-context-protocol}.

\bibitem[Lumer et~al.(2025{\natexlab{a}})Lumer, Subbiah, Burke, Basavaraju, and Huber]{lumericaart25}
Elias Lumer, Vamse Subbiah, James Burke, Pradeep Basavaraju, and Austin Huber.
\newblock Toolshed: Scale tool-equipped agents with advanced rag-tool fusion and tool knowledge bases.
\newblock In \emph{Proceedings of the 17th International Conference on Agents and Artificial Intelligence - Volume 3: ICAART}, pages 1180--1191. INSTICC, SciTePress, 2025{\natexlab{a}}.
\newblock ISBN 978-989-758-737-5.
\newblock \doi{10.5220/0013303000003890}.

\bibitem[Chen et~al.(2024)Chen, Yoon, Sachan, Wang, Cohen-Addad, Bateni, Lee, and Pfister]{chen2024reinvoketoolinvocationrewriting}
Yanfei Chen, Jinsung Yoon, Devendra~Singh Sachan, Qingze Wang, Vincent Cohen-Addad, Mohammadhossein Bateni, Chen-Yu Lee, and Tomas Pfister.
\newblock Re-invoke: Tool invocation rewriting for zero-shot tool retrieval, 2024.
\newblock URL \url{https://arxiv.org/abs/2408.01875}.

\bibitem[Radosevich and Halloran(2025)]{radosevich2025mcpsafetyauditllms}
Brandon Radosevich and John Halloran.
\newblock Mcp safety audit: Llms with the model context protocol allow major security exploits, 2025.
\newblock URL \url{https://arxiv.org/abs/2504.03767}.

\bibitem[Hou et~al.(2025)Hou, Zhao, Wang, and Wang]{hou2025modelcontextprotocolmcp}
Xinyi Hou, Yanjie Zhao, Shenao Wang, and Haoyu Wang.
\newblock Model context protocol (mcp): Landscape, security threats, and future research directions, 2025.
\newblock URL \url{https://arxiv.org/abs/2503.23278}.

\bibitem[{Volo Builds}(2025)]{volobuilds_mcp2025}
{Volo Builds}.
\newblock Mcp has a big problem.
\newblock YouTube video, Apr 2025.
\newblock URL \url{https://www.youtube.com/watch?v=EEE-l41_VQ0}.

\bibitem[{OpenAI}(2025)]{openai_2025}
{OpenAI}.
\newblock Openai.
\newblock \url{https://openai.com/}, 2025.

\bibitem[{Google}(2025)]{google_gemini_2025}
{Google}.
\newblock Gemini.
\newblock \url{https://gemini.google.com/}, 2025.

\bibitem[{Cursor}(2025)]{cursor_2025}
{Cursor}.
\newblock Cursor – the ai code editor.
\newblock \url{https://www.cursor.com/}, 2025.

\bibitem[{Cline}(2025)]{cline_2025}
{Cline}.
\newblock Cline – ai autonomous coding agent for vs code.
\newblock \url{https://cline.bot/}, 2025.

\bibitem[OpenAI(2024)]{openaifunctioncalling2024}
OpenAI.
\newblock Function calling, 2024.
\newblock URL \url{https://platform.openai.com/docs/guides/function-calling}.

\bibitem[Lumer et~al.(2025{\natexlab{b}})Lumer, Basavaraju, Mason, Burke, and Subbiah]{lumer2025graphragtoolfusion}
Elias Lumer, Pradeep~Honaganahalli Basavaraju, Myles Mason, James~A. Burke, and Vamse~Kumar Subbiah.
\newblock Graph rag-tool fusion, 2025{\natexlab{b}}.
\newblock URL \url{https://arxiv.org/abs/2502.07223}.

\bibitem[Peng et~al.(2024)Peng, Zhu, Liu, Bo, Shi, Hong, Zhang, and Tang]{peng2024graphretrievalaugmentedgenerationsurvey}
Boci Peng, Yun Zhu, Yongchao Liu, Xiaohe Bo, Haizhou Shi, Chuntao Hong, Yan Zhang, and Siliang Tang.
\newblock Graph retrieval-augmented generation: A survey, 2024.
\newblock URL \url{https://arxiv.org/abs/2408.08921}.

\bibitem[Singh et~al.(2025)Singh, Ehtesham, Kumar, and Khoei]{singh2025agenticretrievalaugmentedgenerationsurvey}
Aditi Singh, Abul Ehtesham, Saket Kumar, and Tala~Talaei Khoei.
\newblock Agentic retrieval-augmented generation: A survey on agentic rag, 2025.
\newblock URL \url{https://arxiv.org/abs/2501.09136}.

\bibitem[Li et~al.(2023)Li, Zhao, Yu, Song, Li, Yu, Li, Huang, and Li]{li2023apibankcomprehensivebenchmarktoolaugmented}
Minghao Li, Yingxiu Zhao, Bowen Yu, Feifan Song, Hangyu Li, Haiyang Yu, Zhoujun Li, Fei Huang, and Yongbin Li.
\newblock Api-bank: A comprehensive benchmark for tool-augmented llms, 2023.
\newblock URL \url{https://arxiv.org/abs/2304.08244}.

\bibitem[Du et~al.(2024)Du, Wei, and Zhang]{du2024anytoolselfreflectivehierarchicalagents}
Yu~Du, Fangyun Wei, and Hongyang Zhang.
\newblock Anytool: Self-reflective, hierarchical agents for large-scale api calls, 2024.
\newblock URL \url{https://arxiv.org/abs/2402.04253}.

\bibitem[Lumer et~al.(2024)Lumer, Subbiah, Burke, Basavaraju, and Huber]{lumer2024toolshedscaletoolequippedagents}
Elias Lumer, Vamse~Kumar Subbiah, James~A. Burke, Pradeep~Honaganahalli Basavaraju, and Austin Huber.
\newblock Toolshed: Scale tool-equipped agents with advanced rag-tool fusion and tool knowledge bases, 2024.
\newblock URL \url{https://arxiv.org/abs/2410.14594}.

\bibitem[Wu et~al.(2024)Wu, Zhu, Han, Tan, Zhang, and Chen]{wu2024sealtoolsselfinstructtoollearning}
Mengsong Wu, Tong Zhu, Han Han, Chuanyuan Tan, Xiang Zhang, and Wenliang Chen.
\newblock Seal-tools: Self-instruct tool learning dataset for agent tuning and detailed benchmark, 2024.
\newblock URL \url{https://arxiv.org/abs/2405.08355}.

\bibitem[Qin et~al.(2023)Qin, Liang, Ye, Zhu, Yan, Lu, Lin, Cong, Tang, Qian, Zhao, Hong, Tian, Xie, Zhou, Gerstein, Li, Liu, and Sun]{qin2023toolllmfacilitatinglargelanguage}
Yujia Qin, Shihao Liang, Yining Ye, Kunlun Zhu, Lan Yan, Yaxi Lu, Yankai Lin, Xin Cong, Xiangru Tang, Bill Qian, Sihan Zhao, Lauren Hong, Runchu Tian, Ruobing Xie, Jie Zhou, Mark Gerstein, Dahai Li, Zhiyuan Liu, and Maosong Sun.
\newblock Toolllm: Facilitating large language models to master 16000+ real-world apis, 2023.
\newblock URL \url{https://arxiv.org/abs/2307.16789}.

\bibitem[Anantha et~al.(2023)Anantha, Bandyopadhyay, Kashi, Mahinder, Hill, and Chappidi]{anantha2023protipprogressivetoolretrieval}
Raviteja Anantha, Bortik Bandyopadhyay, Anirudh Kashi, Sayantan Mahinder, Andrew~W Hill, and Srinivas Chappidi.
\newblock Protip: Progressive tool retrieval improves planning, 2023.
\newblock URL \url{https://arxiv.org/abs/2312.10332}.

\bibitem[Yuan et~al.(2024{\natexlab{a}})Yuan, Chen, Wang, Fung, Peng, and Ji]{yuan2024craftcustomizingllmscreating}
Lifan Yuan, Yangyi Chen, Xingyao Wang, Yi~R. Fung, Hao Peng, and Heng Ji.
\newblock Craft: Customizing llms by creating and retrieving from specialized toolsets, 2024{\natexlab{a}}.
\newblock URL \url{https://arxiv.org/abs/2309.17428}.

\bibitem[Zheng et~al.(2024)Zheng, Li, Liu, Liu, Luan, and Wang]{zheng2024toolrerankadaptivehierarchyawarereranking}
Yuanhang Zheng, Peng Li, Wei Liu, Yang Liu, Jian Luan, and Bin Wang.
\newblock Toolrerank: Adaptive and hierarchy-aware reranking for tool retrieval, 2024.
\newblock URL \url{https://arxiv.org/abs/2403.06551}.

\bibitem[Robertson and Zaragoza(2009)]{robertson_probabilistic_2009}
Stephen Robertson and Hugo Zaragoza.
\newblock The {Probabilistic} {Relevance} {Framework}: {BM25} and {Beyond}, 2009.
\newblock URL \url{http://dx.doi.org/10.1561/1500000019}.
\newblock ISSN: 1554-0669 Issue: 4 Pages: 333-389 Publication Title: Foundations and Trends® in Information Retrieval Volume: 3.

\bibitem[Gao et~al.(2024)Gao, Xiong, Gao, Jia, Pan, Bi, Dai, Sun, Wang, and Wang]{gao2024retrievalaugmentedgenerationlargelanguage}
Yunfan Gao, Yun Xiong, Xinyu Gao, Kangxiang Jia, Jinliu Pan, Yuxi Bi, Yi~Dai, Jiawei Sun, Meng Wang, and Haofen Wang.
\newblock Retrieval-augmented generation for large language models: A survey, 2024.
\newblock URL \url{https://arxiv.org/abs/2312.10997}.

\bibitem[Hao et~al.(2024{\natexlab{a}})Hao, Liu, Wang, and Hu]{hao2024toolkengptaugmentingfrozenlanguage}
Shibo Hao, Tianyang Liu, Zhen Wang, and Zhiting Hu.
\newblock Toolkengpt: Augmenting frozen language models with massive tools via tool embeddings, 2024{\natexlab{a}}.
\newblock URL \url{https://arxiv.org/abs/2305.11554}.

\bibitem[Patil et~al.(2023)Patil, Zhang, Wang, and Gonzalez]{patil2023gorillalargelanguagemodel}
Shishir~G. Patil, Tianjun Zhang, Xin Wang, and Joseph~E. Gonzalez.
\newblock Gorilla: Large language model connected with massive apis, 2023.
\newblock URL \url{https://arxiv.org/abs/2305.15334}.

\bibitem[Hao et~al.(2024{\natexlab{b}})Hao, Cao, Jin, Liao, Chen, Liu, and Zhao]{hao2024citienhancingtoolutilizing}
Yupu Hao, Pengfei Cao, Zhuoran Jin, Huanxuan Liao, Yubo Chen, Kang Liu, and Jun Zhao.
\newblock Citi: Enhancing tool utilizing ability in large language models without sacrificing general performance, 2024{\natexlab{b}}.
\newblock URL \url{https://arxiv.org/abs/2409.13202}.

\bibitem[Zhu et~al.(2025)Zhu, Shi, Shi, Ren, Wang, Yan, and Yin]{zhu2025dividethenaggregateefficienttoollearning}
Dongsheng Zhu, Weixian Shi, Zhengliang Shi, Zhaochun Ren, Shuaiqiang Wang, Lingyong Yan, and Dawei Yin.
\newblock Divide-then-aggregate: An efficient tool learning method via parallel tool invocation, 2025.
\newblock URL \url{https://arxiv.org/abs/2501.12432}.

\bibitem[Liu et~al.(2024)Liu, Huang, Zeng, Hao, Yu, Li, Wang, Gan, Liu, Yu, Wang, Wang, Ning, Hou, Wang, Wu, Wang, Liu, Wang, Tang, Tu, Shang, Jiang, Tang, Lian, Liu, and Chen]{liu2024toolacewinningpointsllm}
Weiwen Liu, Xu~Huang, Xingshan Zeng, Xinlong Hao, Shuai Yu, Dexun Li, Shuai Wang, Weinan Gan, Zhengying Liu, Yuanqing Yu, Zezhong Wang, Yuxian Wang, Wu~Ning, Yutai Hou, Bin Wang, Chuhan Wu, Xinzhi Wang, Yong Liu, Yasheng Wang, Duyu Tang, Dandan Tu, Lifeng Shang, Xin Jiang, Ruiming Tang, Defu Lian, Qun Liu, and Enhong Chen.
\newblock Toolace: Winning the points of llm function calling, 2024.
\newblock URL \url{https://arxiv.org/abs/2409.00920}.

\bibitem[Zhuang et~al.(2025)Zhuang, Yang, Jiang, Liu, Cheng, Lokegaonkar, Gao, Ping, Liu, Huang, Li, Wang, Chen, Wang, Zhang, Zalmout, Nigam, Yin, and Zhang]{zhuang2025hephaestusimprovingfundamentalagent}
Yuchen Zhuang, Jingfeng Yang, Haoming Jiang, Xin Liu, Kewei Cheng, Sanket Lokegaonkar, Yifan Gao, Qing Ping, Tianyi Liu, Binxuan Huang, Zheng Li, Zhengyang Wang, Pei Chen, Ruijie Wang, Rongzhi Zhang, Nasser Zalmout, Priyanka Nigam, Bing Yin, and Chao Zhang.
\newblock Hephaestus: Improving fundamental agent capabilities of large language models through continual pre-training, 2025.
\newblock URL \url{https://arxiv.org/abs/2502.06589}.

\bibitem[{Anthropic}(2025)]{anthropic_2025}
{Anthropic}.
\newblock Anthropic.
\newblock \url{https://www.anthropic.com/}, 2025.

\bibitem[{Meta Platforms}(2025)]{meta_llama_2025}
{Meta Platforms}.
\newblock Meta llama.
\newblock \url{https://llama.meta.com/}, 2025.

\bibitem[Yuan et~al.(2024{\natexlab{b}})Yuan, Song, Chen, Tan, Shen, Kan, Li, and Yang]{yuan2024easytoolenhancingllmbasedagents}
Siyu Yuan, Kaitao Song, Jiangjie Chen, Xu~Tan, Yongliang Shen, Ren Kan, Dongsheng Li, and Deqing Yang.
\newblock Easytool: Enhancing llm-based agents with concise tool instruction, 2024{\natexlab{b}}.
\newblock URL \url{https://arxiv.org/abs/2401.06201}.

\bibitem[Investopedia(2025)]{investopediafortune10002025}
Investopedia.
\newblock Fortune 1000: Annual list of largest american companies, 2025.
\newblock URL \url{https://www.investopedia.com/terms/f/fortune-1000.asp}.

\bibitem[Aroussi(2025)]{aroussi2025yfinance}
Ran Aroussi.
\newblock {yfinance}: Download market data from yahoo finance api, 2025.
\newblock URL \url{https://yfinance-python.org/}.
\newblock Accessed: 2025-05-02.

\bibitem[jlowin(2025)]{jlowinfastmcp2025}
jlowin.
\newblock Fastmcp: The fast, pythonic way to build mcp servers and clients, 2025.
\newblock URL \url{https://github.com/jlowin/fastmcp}.

\bibitem[{Confident AI}(2025)]{deepeval2025}
{Confident AI}.
\newblock Deepeval: The open-source llm evaluation framework, 2025.
\newblock URL \url{https://www.deepeval.com/}.

\bibitem[Anthropic(2025)]{anthropicthinktool2025}
Anthropic.
\newblock The "think" tool: Enabling claude to stop and think in complex tool use situations, 2025.
\newblock URL \url{https://www.anthropic.com/engineering/claude-think-tool}.

\bibitem[Surapaneni et~al.(2025)Surapaneni, Jha, Vakoc, and Segal]{googlea2a2025}
Rao Surapaneni, Miku Jha, Michael Vakoc, and Todd Segal.
\newblock Announcing the agent2agent protocol (a2a), 2025.
\newblock URL \url{https://developers.googleblog.com/en/a2a-a-new-era-of-agent-interoperability/}.

\end{thebibliography}

\appendix
\section*{Appendix}

\renewcommand{\thetable}{A\arabic{table}}  

\subsection*{Full Retrieval Performance Table from Experiment 1}

Table~\ref{tab:retrieval_performance_2} provides the complete retrieval metrics from Experiment 1 using the concatenation storage strategy.

\begingroup
\scriptsize
\begin{longtable}{@{}p{3.6cm}c*{9}{>{\centering\arraybackslash}p{0.8cm}}@{}}
\caption{Retrieval performance at $K=5$ and metric using the \textbf{concatenation} storage strategy.}%
\label{tab:retrieval_performance_2}\\
\toprule
\textbf{Embedding Model} & \textbf{SQ} &
    \multicolumn{3}{c}{\textbf{\boldmath$K=1$}} &
    \multicolumn{3}{c}{\textbf{\boldmath$K=5$}} &
    \multicolumn{3}{c}{\textbf{\boldmath$K=10$}} \\

    \cmidrule(lr){3-5}\cmidrule(lr){6-8}\cmidrule(lr){9-11}
    & & \textbf{NDCG} & \textbf{Recall} & \textbf{MAP} & \textbf{NDCG} & \textbf{Recall} & \textbf{MAP} & \textbf{NDCG} & \textbf{Recall} & \textbf{MAP} \\

\midrule
\endfirsthead

\multicolumn{11}{c}{\tablename\ \thetable{} -- continued from previous page}\\
\toprule
\textbf{Embedding Model} & \textbf{SQ} &
    \multicolumn{3}{c}{\textbf{\boldmath$K=1$}} &
    \multicolumn{3}{c}{\textbf{\boldmath$K=5$}} &
    \multicolumn{3}{c}{\textbf{\boldmath$K=10$}} \\
    \cmidrule(lr){3-5}\cmidrule(lr){6-8}\cmidrule(lr){9-11}
    & & \textbf{NDCG} & \textbf{Recall} & \textbf{MAP} & \textbf{NDCG} & \textbf{Recall} & \textbf{MAP} & \textbf{NDCG} & \textbf{Recall} & \textbf{MAP} \\

\midrule
\endhead

\midrule
\multicolumn{11}{r}{\textit{Continued on next page}}\\
\midrule
\endfoot

\bottomrule
\endlastfoot

\multicolumn{11}{@{}l}{\textit{Vector Search}}\\
\multirow{3}{3.6cm}{OpenAI text-embedding-3-large}
 &  0 & 0.88 & 0.49 & 0.49 & 0.62 & 0.83 & 0.50 & 0.43 & 0.88 & 0.25 \\
 &  5 & \underline{0.94} & 0.51 & 0.51 & 0.59 & 0.85 & 0.47 & 0.44 & 0.88 & 0.27 \\
  & 10 & \underline{0.94} & \underline{0.52} & \underline{0.52} & 0.63 & 0.91 & 0.50 & 0.45 & \underline{0.93} & 0.26 \\
\addlinespace
\multirow{3}{3.6cm}{OpenAI text-embedding-3-small}
 &  0 & 0.84 & 0.48 & 0.48 & 0.64 & 0.82 & 0.53 & 0.43 & 0.84 & 0.27 \\
 &  5 & 0.90 & 0.50 & 0.50 & 0.65 & 0.84 & 0.54 & 0.43 & 0.85 & 0.27 \\
 & 10 & 0.92 & \underline{0.52} & \underline{0.52} & 0.64 & 0.85 & 0.53 & 0.45 & 0.90 & 0.28 \\
\addlinespace
\multirow{3}{3.6cm}{OpenAI text-embedding-ada-002}
 &  0 & 0.86 & 0.48 & 0.48 & 0.60 & 0.81 & 0.50 & 0.40 & 0.83 & 0.24 \\
 &  5 & 0.86 & 0.48 & 0.48 & 0.59 & 0.83 & 0.46 & 0.40 & 0.86 & 0.23 \\
 & 10 & \underline{0.94} & \textbf{0.53} & \textbf{0.53} & 0.59 & 0.85 & 0.47 & 0.43 & 0.87 & 0.25 \\
\addlinespace
\multirow{3}{3.6cm}{Amazon titan-embed-text-v1}
 &  0 & 0.82 & 0.45 & 0.45 & 0.62 & 0.79 & 0.51 & 0.40 & 0.80 & 0.24 \\
 &  5 & 0.86 & 0.48 & 0.48 & 0.59 & 0.80 & 0.48 & 0.43 & 0.83 & 0.27 \\
 & 10 & 0.86 & 0.48 & 0.48 & 0.57 & 0.79 & 0.47 & 0.41 & 0.83 & 0.25 \\
\addlinespace
\multirow{3}{3.6cm}{VertexAI text-embedding-005}
 &  0 & 0.92 & 0.51 & 0.51 & 0.63 & 0.87 & 0.51 & 0.46 & 0.88 & 0.30 \\
 &  5 & \underline{0.94} & 0.51 & 0.51 & 0.64 & 0.87 & 0.51 & 0.46 & 0.92 & 0.28 \\
 & 10 & \textbf{0.96} & \textbf{0.53} & \textbf{0.53} & 0.65 & 0.91 & 0.51 & \textbf{0.50} & \textbf{0.95} & \textbf{0.32} \\

\midrule
\multicolumn{11}{@{}l}{\textit{Text Search (BM25)}}\\
\multirow{3}{3.6cm}{OpenAI text-embedding-3-large}
 &  0 & 0.52 & 0.31 & 0.31 & 0.43 & 0.57 & 0.33 & 0.33 & 0.63 & 0.21 \\
 &  5 & 0.62 & 0.36 & 0.36 & 0.47 & 0.63 & 0.38 & 0.33 & 0.66 & 0.19 \\
 & 10 & 0.58 & 0.34 & 0.34 & 0.48 & 0.65 & 0.39 & 0.35 & 0.68 & 0.22 \\
\addlinespace
\multirow{3}{3.6cm}{OpenAI text-embedding-3-small}
 &  0 & 0.58 & 0.34 & 0.34 & 0.41 & 0.57 & 0.32 & 0.32 & 0.63 & 0.20 \\
 &  5 & 0.60 & 0.35 & 0.35 & 0.48 & 0.64 & 0.40 & 0.35 & 0.69 & 0.21 \\
 & 10 & 0.64 & 0.39 & 0.39 & 0.52 & 0.68 & 0.43 & 0.35 & 0.70 & 0.22 \\
\addlinespace
\multirow{3}{3.6cm}{OpenAI text-embedding-ada-002}
 &  0 & 0.56 & 0.32 & 0.32 & 0.42 & 0.59 & 0.33 & 0.32 & 0.64 & 0.19 \\
 &  5 & 0.62 & 0.35 & 0.35 & 0.48 & 0.64 & 0.39 & 0.37 & 0.72 & 0.23 \\
 & 10 & 0.64 & 0.39 & 0.39 & 0.48 & 0.66 & 0.39 & 0.34 & 0.69 & 0.20 \\
\addlinespace
\multirow{3}{3.6cm}{Amazon titan-embed-text-v1}
 &  0 & 0.54 & 0.32 & 0.32 & 0.40 & 0.56 & 0.31 & 0.31 & 0.62 & 0.18 \\
 &  5 & 0.66 & 0.38 & 0.38 & 0.48 & 0.65 & 0.37 & 0.34 & 0.67 & 0.21 \\
 & 10 & 0.64 & 0.38 & 0.38 & 0.49 & 0.66 & 0.39 & 0.34 & 0.69 & 0.21 \\
\addlinespace
\multirow{3}{3.6cm}{VertexAI text-embedding-005}
 &  0 & 0.54 & 0.33 & 0.33 & 0.42 & 0.58 & 0.32 & 0.33 & 0.63 & 0.21 \\
 &  5 & 0.56 & 0.32 & 0.32 & 0.49 & 0.65 & 0.40 & 0.35 & 0.69 & 0.22 \\
 & 10 & 0.64 & 0.38 & 0.38 & 0.52 & 0.66 & 0.44 & 0.36 & 0.68 & 0.23 \\

\midrule
\multicolumn{11}{@{}l}{\textit{Hybrid Search}}\\
\multirow{3}{3.6cm}{OpenAI text-embedding-3-large}
 &  0 & 0.70 & 0.39 & 0.39 & 0.53 & 0.72 & 0.42 & 0.38 & 0.75 & 0.23 \\
 &  5 & 0.76 & 0.42 & 0.42 & 0.57 & 0.76 & 0.45 & 0.43 & 0.83 & 0.26 \\
 & 10 & 0.82 & 0.45 & 0.45 & 0.63 & 0.84 & 0.50 & 0.46 & 0.87 & 0.29 \\
\addlinespace
\multirow{3}{3.6cm}{OpenAI text-embedding-3-small}
 &  0 & 0.72 & 0.40 & 0.40 & 0.55 & 0.74 & 0.45 & 0.38 & 0.74 & 0.24 \\
 &  5 & 0.76 & 0.43 & 0.43 & 0.62 & 0.78 & 0.50 & 0.44 & 0.84 & 0.27 \\ 
 & 10 & 0.76 & 0.44 & 0.44 & 0.64 & 0.83 & 0.53 & 0.43 & 0.84 & 0.26 \\
\addlinespace
\multirow{3}{3.6cm}{OpenAI text-embedding-ada-002}
 &  0 & 0.72 & 0.39 & 0.39 & 0.58 & 0.76 & 0.47 & 0.41 & 0.79 & 0.26 \\
 &  5 & 0.74 & 0.41 & 0.41 & 0.60 & 0.82 & 0.49 & 0.42 & 0.83 & 0.26 \\
 & 10 & 0.74 & 0.44 & 0.44 & 0.57 & 0.77 & 0.45 & 0.41 & 0.78 & 0.26 \\
\addlinespace
\multirow{3}{3.6cm}{Amazon titan-embed-text-v1}
 &  0 & 0.80 & 0.45 & 0.45 & 0.54 & 0.72 & 0.43 & 0.38 & 0.74 & 0.24 \\
 &  5 & 0.76 & 0.42 & 0.42 & 0.60 & 0.76 & 0.49 & 0.41 & 0.78 & 0.25 \\
 & 10 & 0.82 & 0.46 & 0.46 & 0.59 & 0.76 & 0.47 & 0.41 & 0.78 & 0.25 \\
\addlinespace
\multirow{3}{3.6cm}{VertexAI text-embedding-005}
 &  0 & 0.74 & 0.41 & 0.41 & 0.58 & 0.75 & 0.47 & 0.39 & 0.78 & 0.24 \\
 &  5 & 0.80 & 0.45 & 0.45 & 0.59 & 0.78 & 0.47 & 0.39 & 0.79 & 0.23 \\
 & 10 & 0.80 & 0.44 & 0.44 & 0.63 & 0.83 & 0.51 & 0.44 & 0.86 & 0.28 \\

\midrule
\addlinespace
\multicolumn{11}{@{}l}{\textit{Cohere Reranking (v3-english)}}\\

\multirow{3}{3.6cm}{OpenAI text-embedding-3-large}
 &  0 & 0.84 & 0.46 & 0.46 & 0.64 & 0.85 & 0.52 & 0.43 & 0.88 & 0.25 \\
 &  5 & 0.86 & 0.47 & 0.47 & 0.61 & 0.85 & 0.49 & 0.44 & 0.88 & 0.27 \\
 & 10 & 0.86 & 0.47 & 0.47 & 0.64 & 0.90 & 0.51 & 0.45 & 0.93 & 0.26 \\
\addlinespace
\multirow{3}{3.6cm}{OpenAI text-embedding-3-small}
 &  0 & 0.84 & 0.46 & 0.46 & 0.63 & 0.83 & 0.52 & 0.43 & 0.84 & 0.27 \\
 &  5 & 0.86 & 0.47 & 0.47 & 0.65 & 0.85 & 0.54 & 0.43 & 0.85 & 0.27 \\
 & 10 & 0.86 & 0.47 & 0.47 & 0.64 & 0.83 & 0.53 & 0.45 & 0.90 & 0.28 \\
\addlinespace
\multirow{3}{3.6cm}{OpenAI text-embedding-ada-002}
 &  0 & 0.86 & 0.47 & 0.47 & 0.63 & 0.83 & 0.52 & 0.40 & 0.83 & 0.24 \\
 &  5 & 0.88 & 0.48 & 0.48 & 0.58 & 0.83 & 0.45 & 0.40 & 0.86 & 0.23 \\
 & 10 & 0.86 & 0.47 & 0.47 & 0.61 & 0.83 & 0.48 & 0.43 & 0.87 & 0.25 \\
\addlinespace
\multirow{3}{3.6cm}{Amazon titan-embed-text-v1}
 &  0 & 0.84 & 0.46 & 0.46 & 0.62 & 0.80 & 0.52 & 0.40 & 0.80 & 0.24 \\
 &  5 & 0.86 & 0.47 & 0.47 & 0.60 & 0.81 & 0.49 & 0.43 & 0.83 & 0.27 \\
 & 10 & 0.84 & 0.46 & 0.46 & 0.60 & 0.81 & 0.49 & 0.41 & 0.83 & 0.25 \\
\addlinespace
\multirow{3}{3.6cm}{VertexAI text-embedding-005}
 &  0 & 0.86 & 0.47 & 0.47 & 0.65 & 0.87 & 0.53 & 0.46 & 0.88 & 0.30 \\
 &  5 & 0.86 & 0.47 & 0.47 & 0.64 & 0.88 & 0.52 & 0.46 & 0.92 & 0.28 \\
 & 10 & 0.88 & 0.47 & 0.47 & 0.67 & 0.90 & 0.54 & \textbf{0.50} & \textbf{0.95} & \textbf{0.32} \\

\midrule
\multicolumn{11}{@{}l}{\textit{LLM Reranking (GPT-4o)}}\\
\multirow{3}{3.6cm}{OpenAI text-embedding-3-large}
 &  0 & 0.86 & 0.47 & 0.47 & 0.66 & 0.88 & 0.54 & 0.43 & 0.88 & 0.25 \\
 &  5 & 0.86 & 0.47 & 0.47 & 0.64 & 0.88 & 0.51 & 0.44 & 0.88 & 0.27 \\
 & 10 & 0.88 & 0.49 & 0.49 & 0.67 & 0.91 & 0.54 & 0.46 & \underline{0.93} & 0.28 \\
\addlinespace
\multirow{3}{3.6cm}{OpenAI text-embedding-3-small}
 &  0 & 0.86 & 0.46 & 0.46 & 0.64 & 0.84 & 0.53 & 0.43 & 0.84 & 0.27 \\  
 &  5 & 0.86 & 0.47 & 0.47 & 0.65 & 0.85 & 0.54 & 0.43 & 0.85 & 0.27 \\
 & 10 & 0.88 & 0.49 & 0.49 & 0.67 & 0.87 & 0.56 & 0.44 & 0.88 & 0.28 \\
\addlinespace
\multirow{3}{3.6cm}{OpenAI text-embedding-ada-002}
 &  0 & 0.90 & 0.49 & 0.49 & 0.61 & 0.83 & 0.50 & 0.41 & 0.83 & 0.25 \\
 &  5 & 0.88 & 0.49 & 0.49 & 0.62 & 0.86 & 0.50 & 0.40 & 0.86 & 0.23 \\
 & 10 & 0.92 & \textbf{0.53} & \textbf{0.53} & 0.61 & 0.87 & 0.47 & 0.44 & 0.87 & 0.27 \\
\addlinespace
\multirow{3}{3.6cm}{Amazon titan-embed-text-v1}
 &  0 & 0.86 & 0.47 & 0.47 & 0.61 & 0.80 & 0.50 & 0.41 & 0.80 & 0.25 \\
 &  5 & 0.90 & 0.50 & 0.50 & 0.62 & 0.83 & 0.51 & 0.44 & 0.83 & 0.28 \\
 & 10 & 0.88 & 0.49 & 0.49 & 0.61 & 0.83 & 0.51 & 0.42 & 0.83 & 0.26 \\
\addlinespace
\multirow{3}{3.6cm}{VertexAI text-embedding-005}
 &  0 & 0.88 & 0.49 & 0.49 & 0.67 & 0.87 & 0.56 & \underline{0.47} & 0.88 & \underline{0.31} \\
 &  5 & 0.88 & 0.49 & 0.49 & 0.68 & 0.90 & 0.55 & 0.46 & 0.92 & 0.28 \\
 & 10 & 0.92 & 0.51 & 0.51 & 0.70 & \textbf{0.94} & \underline{0.58} & \textbf{0.50} & \textbf{0.95} & \textbf{0.32} \\

\midrule
\multicolumn{11}{@{}l}{\textit{LLM Reranking (Claude Sonnet 3.7)}}\\
\multirow{3}{3.6cm}{OpenAI text-embedding-3-large}
 &  0 & 0.90 & 0.51 & 0.51 & 0.66 & 0.88 & 0.54 & 0.43 & 0.88 & 0.25 \\
 &  5 & 0.92 & 0.51 & 0.51 & 0.63 & 0.88 & 0.50 & 0.44 & 0.88 & 0.27 \\
 & 10 & \underline{0.94} & \textbf{0.53} & \textbf{0.53} & 0.67 & 0.91 & 0.54 & 0.45 & \underline{0.93} & 0.26 \\
\addlinespace
\multirow{3}{3.6cm}{OpenAI text-embedding-3-small}
 &  0 & 0.92 & 0.51 & 0.51 & 0.66 & 0.84 & 0.55 & 0.43 & 0.84 & 0.27 \\
 &  5 & 0.90 & 0.51 & 0.51 & 0.66 & 0.85 & 0.54 & 0.43 & 0.85 & 0.27 \\
 & 10 & 0.92 & \textbf{0.53} & \textbf{0.53} & \underline{0.70} & 0.89 & \textbf{0.59} & 0.45 & 0.90 & 0.28 \\
\addlinespace
\multirow{3}{3.6cm}{OpenAI text-embedding-ada-002}
 &  0 & 0.92 & 0.51 & 0.51 & 0.62 & 0.83 & 0.52 & 0.40 & 0.83 & 0.24 \\
 &  5 & 0.90 & 0.51 & 0.51 & 0.61 & 0.86 & 0.48 & 0.40 & 0.86 & 0.23 \\
 & 10 & 0.92 & \textbf{0.53} & \textbf{0.53} & 0.61 & 0.87 & 0.48 & 0.43 & 0.87 & 0.25 \\
\addlinespace
\multirow{3}{3.6cm}{Amazon titan-embed-text-v1}
 &  0 & 0.86 & 0.47 & 0.47 & 0.63 & 0.80 & 0.53 & 0.40 & 0.80 & 0.24 \\
 &  5 & 0.88 & 0.49 & 0.49 & 0.62 & 0.83 & 0.51 & 0.43 & 0.83 & 0.27 \\
 & 10 & 0.90 & 0.51 & 0.51 & 0.63 & 0.83 & 0.52 & 0.41 & 0.83 & 0.25 \\
\addlinespace
\multirow{3}{3.6cm}{VertexAI text-embedding-005}
 &  0 & 0.92 & \underline{0.52} & 0.52 & 0.67 & 0.87 & 0.55 & 0.46 & 0.88 & 0.30 \\
 &  5 & 0.92 & 0.51 & 0.51 & 0.69 & 0.89 & \underline{0.58} & 0.46 & 0.92 & 0.28 \\
 & 10 & \underline{0.94} & \textbf{0.53} & \textbf{0.53} & \textbf{0.71} & \underline{0.93} & \textbf{0.59} & \textbf{0.50} & \textbf{0.95} & \textbf{0.32} \\

\end{longtable}
\endgroup

\subsection*{LLM Agent Evaluation (Experiment 2)}

\begingroup
\scriptsize
\begin{longtable}{%
    >{\raggedright\arraybackslash}p{2.6cm} 
    >{\raggedright\arraybackslash}p{4.0cm} 
    >{\centering\arraybackslash}p{3.2cm}   
    >{\centering\arraybackslash}p{3.2cm}}  
\caption{Agent-level Tool Correctness and Task Completion rates at $k=5$ using the concatenation strategy and 10 questions per tool. "Vector Search + Cohere Reranker" uses a trained reranker.}%
\label{tab:agent_performance_2}\\
\toprule
\textbf{LLM Agent} & \textbf{Retrieval Method} & \textbf{Tool Correctness (\%)} & \textbf{Task Completion (\%)} \\
\midrule
\endfirsthead

\multicolumn{4}{c}{\tablename\ \thetable{} -- continued from previous page}\\
\toprule
\textbf{LLM Agent} & \textbf{Retrieval Method} & \textbf{Tool Correctness (\%)} & \textbf{Task Completion (\%)} \\
\midrule
\endhead

\midrule
\multicolumn{4}{r}{\textit{Continued on next page}}\\
\midrule
\endfoot

\bottomrule
\endlastfoot
GPT 4.1 & Text Search & 39.1 & 73.2 \\
GPT 4.1 & Vector Search & 45.0 & 82.4 \\
GPT 4.1 & Vector Search + Cohere Reranker & 44.8 & 85.8 \\
\midrule
GPT 4.1-mini & Text Search & 48.5 & 74.4 \\
GPT 4.1-mini & Vector Search & \underline{53.0} & 85.5 \\
GPT 4.1-mini & Vector Search + Cohere Reranker & 52.8 & 81.1 \\
\midrule
GPT 4.1-nano & Text Search & 31.1 & 75.6 \\
GPT 4.1-nano & Vector Search & 38.4 & 81.1 \\
GPT 4.1-nano & Vector Search + Cohere Reranker & 31.7 & 80.6 \\
\midrule
GPT 4o & Text Search & 47.9 & 75.9 \\
GPT 4o & Vector Search & 51.7 & 86.5 \\
GPT 4o & Vector Search + Cohere Reranker & 51.7 & 83.9 \\
\midrule
GPT 4o-mini & Text Search & 49.9 & 81.2 \\
GPT 4o-mini & Vector Search & 50.7 & 86.5 \\
GPT 4o-mini & Vector Search + Cohere Reranker & \textbf{54.0} & 86.7 \\
\midrule
GPT o4-mini & Text Search & 37.5 & 84.1 \\
GPT o4-mini & Vector Search & 38.6 & 81.3 \\
GPT o4-mini & Vector Search + Cohere Reranker & 40.7 & 85.6 \\
\midrule
GPT o3 & Text Search & 36.1 & 78.9 \\
GPT o3 & Vector Search & 22.2 & \underline{88.9} \\
GPT o3 & Vector Search + Cohere Reranker & 36.1 & \textbf{94.4} \\
\midrule
Claude 3.7 Sonnet & Text Search & 23.9 & 42.9 \\
Claude 3.7 Sonnet & Vector Search & 28.5 & 73.2 \\
Claude 3.7 Sonnet & Vector Search + Cohere Reranker & 23.1 & 69.4 \\
\midrule
Claude 3.5 Sonnet & Text Search & 13.9 & 45.0 \\
Claude 3.5 Sonnet & Vector Search & 29.2 & 29.6 \\
Claude 3.5 Sonnet & Vector Search + Cohere Reranker & 13.9 & 44.4 \\
\midrule
Llama 3.3 70B & Text Search & 0.0 & 48.0 \\
Llama 3.3 70B & Vector Search & 0.0 & 51.0 \\
Llama 3.3 70B & Vector Search + Cohere Reranker & 0.0 & 48.8 \\

\end{longtable}
\endgroup

\subsection*{Retrieval Performance Across Weighting Strategies (Experiment 3)}

Table~\ref{tab:weighting_strategies} summarizes retrieval performance at $SQ=10$ for vector, text, and reranking-based strategies, comparing the concatenation and TDWA (Tool-Description Weighted Average) approaches with different weight variants. We report results for $K=1$, $K=5$, and $K=10$ using NDCG, Recall, and MAP.

\begingroup
\scriptsize
\begin{longtable}{@{}p{3.6cm}p{1.0cm}p{0.7cm}*{9}{>{\centering\arraybackslash}p{0.75cm}}@{}}
\caption{Retrieval Performance Across Search Strategies ($SQ=10$).}%
\label{tab:weighting_strategies}\\
\toprule
\textbf{Embedding Model} & \textbf{Strategy} & \textbf{Weights} &
\multicolumn{3}{c}{\textbf{\boldmath$K=1$}} & \multicolumn{3}{c}{\textbf{\boldmath$K=5$}} & \multicolumn{3}{c}{\textbf{\boldmath$K=10$}} \\
\cmidrule(lr){4-6}\cmidrule(lr){7-9}\cmidrule(lr){10-12}
& & & \textbf{NDCG} & \textbf{Recall} & \textbf{MAP} & \textbf{NDCG} & \textbf{Recall} & \textbf{MAP} & \textbf{NDCG} & \textbf{Recall} & \textbf{MAP} \\

\midrule
\endfirsthead

\multicolumn{12}{c}{\tablename\ \thetable{} -- continued}\\
\toprule
\textbf{Embedding Model} & \textbf{Strategy} & \textbf{Weights} &
\multicolumn{3}{c}{$K=1$} & \multicolumn{3}{c}{$K=5$} & \multicolumn{3}{c}{$K=10$} \\
\cmidrule(lr){4-6}\cmidrule(lr){7-9}\cmidrule(lr){10-12}
& & & NDCG & Recall & MAP & NDCG & Recall & MAP & NDCG & Recall & MAP \\
\midrule
\endhead

\midrule
\multicolumn{12}{r}{\textit{Continued on next page}}\\
\midrule
\endfoot

\bottomrule
\endlastfoot
\multicolumn{12}{@{}l}{\textit{Vector Search}}\\
OpenAI text-embedding-3-large & Concat & -- & \textbf{0.94} & 0.52 & 0.52 & 0.634 & \textbf{0.912} & 0.499 & \textbf{0.449} & \textbf{0.933} & 0.265 \\
OpenAI text-embedding-3-large & TDWA & var-1 & 0.86 & 0.489 & 0.489 & 0.631 & 0.886 & 0.504 & 0.431 & 0.89 & 0.255 \\
OpenAI text-embedding-3-large & TDWA & var-2 & 0.9 & 0.509 & 0.509 & 0.62 & 0.891 & 0.485 & 0.447 & \underline{0.899} & \underline{0.273} \\
\midrule
\multicolumn{12}{@{}l}{\textit{Text Search (BM25)}}\\
-- & BM25 & -- & 0.64 & 0.382 & 0.382 & 0.492 & 0.674 & 0.396 & 0.354 & 0.691 & 0.220 \\
\midrule
\multicolumn{12}{@{}l}{\textit{Cohere Reranking (v3-english)}}\\
OpenAI text-embedding-3-large & Concat & -- & 0.86 & 0.465 & 0.465 & 0.642 & 0.896 & 0.51 & \textbf{0.449} & \textbf{0.933} & 0.265 \\
OpenAI text-embedding-3-large & TDWA & var-1 & 0.84 & 0.455 & 0.455 & 0.644 & 0.855 & \underline{0.528} & 0.431 & 0.89 & 0.255 \\
OpenAI text-embedding-3-large & TDWA & var-2 & 0.84 & 0.455 & 0.455 & 0.629 & 0.839 & 0.511 & 0.447 & \underline{0.899} & \underline{0.273} \\
\midrule
\multicolumn{12}{@{}l}{\textit{LLM Reranking (GPT-4o)}}\\
OpenAI text-embedding-3-large & Concat & -- & 0.88 & 0.485 & 0.485 & \underline{0.669} & \underline{0.906} & 0.545 & 0.461 & \textbf{0.933} & \textbf{0.281} \\
OpenAI text-embedding-3-large & TDWA & var-1 & \underline{0.92} & 0.515 & 0.515 & 0.638 & 0.885 & 0.505 & 0.429 & 0.885 & 0.255 \\
OpenAI text-embedding-3-large & TDWA & var-2 & 0.88 & 0.485 & 0.485 & 0.656 & 0.889 & \underline{0.528} & 0.447 & \underline{0.899} & \underline{0.273} \\
\midrule
\multicolumn{12}{@{}l}{\textit{LLM Reranking (Claude Sonnet 3.7)}}\\
OpenAI text-embedding-3-large & Concat & -- & \textbf{0.94} & \underline{0.53} & \underline{0.53} & \textbf{0.672} & \textbf{0.912} & \textbf{0.539} & \textbf{0.449} & \textbf{0.933} & 0.265 \\
OpenAI text-embedding-3-large & TDWA & var-1 & \textbf{0.94} & \textbf{0.535} & \textbf{0.535} & 0.638 & 0.885 & 0.508 & 0.431 & 0.89 & 0.255 \\
OpenAI text-embedding-3-large & TDWA & var-2 & \underline{0.92} & 0.525 & 0.525 & 0.644 & 0.889 & 0.511 & \underline{0.447} & \underline{0.899} & \underline{0.273} \\
\midrule
\end{longtable}
\endgroup

\subsection*{Multi-Company Synthetic Queries}

The following complex queries were synthesized to test multi-hop reasoning and comparison across multiple companies. These involve combinations of revenue, stock trends, and analyst metrics.

\begin{itemize}
  \item How do \texttt{\{company 1\}}, \texttt{\{company 2\}}, \texttt{\{company 3\}}, and \texttt{\{company 4\}} compare in terms of stock performance and analyst expectations over the past week?
  \item Can you detail a year-over-year comparison of \texttt{\{company 1\}}'s financial metrics by looking at its 2023 and 2024 revenue and net income?
  \item Analyze the stock performance of \texttt{\{company 1\}}, \texttt{\{company 2\}}, and \texttt{\{company 3\}} by reviewing each company's current stock price, recent weekly price history, and current analyst price targets to determine the best overall performer.
  \item Out of the following 15 companies: \texttt{\{company 1\}}, \texttt{\{company 2\}}, \texttt{\{company 3\}}, \texttt{\{company 4\}}, \texttt{\{company 5\}}, \texttt{\{company 6\}}, \texttt{\{company 7\}}, \texttt{\{company 8\}}, \texttt{\{company 9\}}, \texttt{\{company 10\}}, \texttt{\{company 11\}}, \texttt{\{company 12\}}, \texttt{\{company 13\}}, \texttt{\{company 14\}}, and \texttt{\{company 15\}}, determine which one reported the highest revenue for 2024.
  \item Determine the relative financial strength of \texttt{\{company 1\}} compared to \texttt{\{company 2\}} and \texttt{\{company 3\}} by comparing their 2024 net income margins.
  \item Compare the 2024 revenue of \texttt{\{company 1\}}, \texttt{\{company 2\}}, \texttt{\{company 3\}}, \texttt{\{company 4\}}, and \texttt{\{company 5\}}, and determine which company achieved the highest revenue.
\end{itemize}

\subsection*{System Prompt Used for LLM Agents}

All agents used the same system prompt to guide their behavior and interaction with the tool-based knowledge base. The prompt is shown below:

\begin{quote}
\small
\texttt{You are an intelligent financial assistant. You have access to a large knowledge base of tools.\\
The only way to use the large knowledge base of tools is to use the 'get\_mcp\_servers' tool to search relevant ones.\\
Query the 'get\_mcp\_servers' knowledge base by passing in a query for a tool you want to search for.\\
IF YOU NEED MULTIPLE TOOLS, USE PARALLEL TOOL CALLING, EACH TOOL CALL TO SEARCH FOR SPECIFIC TOOLS.}
\end{quote}

\subsection*{Prompt Used for Generating Synthetic User Query Instances}

The following prompt was used to generate diverse, natural language financial queries paired with their corresponding tool calls. These examples form the foundation of our synthetic dataset used for evaluation and training.

\begin{quote}
\small
You are an advanced AI that generates natural language financial queries along with their structured tool calls, using predefined financial tools.

For each sample, generate:
\begin{enumerate}
    \item A realistic and diverse financial query a user might ask, involving one or more of the following tools. Use \texttt{\{company\}} as a placeholder for the company name.
    \item A corresponding list of \textbf{tool\_calls} that would be needed to answer that query.
    \item Ensure the tool function names are properly formed and arguments are valid.
\end{enumerate}

\textbf{Available Tools:}
\begin{itemize}
    \item \texttt{get\_\{company\}\_current\_stock\_price}: Return the most recent trading price for \texttt{\{company\}}.
    \item \texttt{get\_\{company\}\_stock\_price\_history}: Requires \texttt{"name"} arg — one of \texttt{"d"}, \texttt{"w"}, or \texttt{"m"} for daily, weekly, or monthly.
    \item \texttt{get\_\{company\}\_analyst\_price\_targets}: Requires \texttt{"name"} arg — one of \texttt{"current"}, \texttt{"low"}, \texttt{"high"}, \texttt{"mean"}, or \texttt{"median"}.
    \item \texttt{get\_\{company\}\_revenue}: Optional \texttt{"name"} arg — a year (e.g., \texttt{2024}) or omitted for all years.
    \item \texttt{get\_\{company\}\_net\_income}: Same as above.
\end{itemize}

\textbf{Instructions:}
\begin{itemize}
    \item Vary phrasing styles across queries.
    \item Include a mix of single-tool and multi-tool queries.
    \item Make sure queries feel human-written (natural, concise, and varied).
    \item For tools requiring \texttt{"name"} arguments, include the proper ones in the \texttt{args}.
    \item Use \texttt{\{company\}} exactly as-is in the query text and tool names.
    \item Output 200 examples in the following format:
\end{itemize}
\end{quote}

\begin{verbatim}
{
  "query": "How has {company}'s stock performed on a weekly vs monthly basis?",
  "tool_calls": [
    {"name": "get_{company}_stock_price_history", "args": [{"name": "w"}]},
    {"name": "get_{company}_stock_price_history", "args": [{"name": "m"}]}
  ]
}
\end{verbatim}

\subsection*{Example Synthetic Questions Used for Retrieval Evaluation}

Table~\ref{tab:synthetic_questions} shows sample synthetic questions generated for five representative financial tool functions. These were used to simulate user queries in retrieval experiments.

\begingroup
\scriptsize
\begin{longtable}{p{4.2cm}p{11.5cm}}
\caption{Example synthetic questions used for five financial tool functions.}
\label{tab:synthetic_questions} \\
\toprule
\textbf{Tool Function} & \textbf{Example Synthetic Questions} \\
\midrule
\endfirsthead
\multicolumn{2}{c}{\tablename\ \thetable{} -- continued from previous page} \\
\toprule
\textbf{Tool Function} & \textbf{Example Synthetic Questions} \\
\midrule
\endhead
\bottomrule
\endfoot

\texttt{get\_\{company\}\_current\_stock\_price} &
What is the current stock price of \{company\}?; How much is \{company\}'s stock trading for right now?; Please show me the current market price of \{company\}'s shares.; What's the latest price at which \{company\} stock is trading?; How much is one share of \{company\} worth at the moment? \\
\addlinespace
\texttt{get\_\{company\}\_stock\_price\_history} &
Can you show me the daily closing stock prices for \{company\} over the past year?; What are the last 10 weekly closing prices for \{company\}?; I'd like to see \{company\}'s monthly stock price history for the past year.; Retrieve the recent daily stock price trend for \{company\}; Provide the last 10 monthly closing price points for \{company\}. \\
\addlinespace
\texttt{get\_\{company\}\_analyst\_price\_targets} &
What is the current analyst price target for \{company\}?; Can you fetch the low forecasted price target?; Show me the mean analyst price target for \{company\}; What high target have analysts set for \{company\}?; Provide the median forecasted price. \\
\addlinespace
\texttt{get\_\{company\}\_revenue} &
What is \{company\}'s revenue for the year 2022?; Can you show me annual revenue figures for 2021?; I'd like to see the revenue data for 2020.; Get the revenue details for the latest fiscal year.; Provide \{company\}'s revenue history by year. \\
\addlinespace
\texttt{get\_\{company\}\_net\_income} &
What is \{company\}'s net income for 2022?; Show me net income trends over recent years.; Can you provide net income details for 2020?; Retrieve the most recent net income value.; Please fetch net income for a specific year, e.g., 2023. \\
\end{longtable}
\endgroup

\subsection*{Tool Descriptions}

Table~\ref{tab:tool_descriptions} lists the descriptions for the five tool functions evaluated. These descriptions guided the generation of synthetic questions and informed LLM usage.

\begingroup
\scriptsize
\begin{longtable}{p{3.8cm}p{4.5cm}p{7.4cm}}
\caption{Descriptions and parameters of financial tools used in retrieval and generation experiments.}
\label{tab:tool_descriptions} \\
\toprule
\textbf{Tool Name} & \textbf{Parameters} & \textbf{Description} \\
\midrule
\endfirsthead
\multicolumn{3}{c}{\tablename\ \thetable{} -- continued from previous page} \\
\toprule
\textbf{Tool Name} & \textbf{Parameters} & \textbf{Description} \\
\midrule
\endhead
\bottomrule
\endfoot

\texttt{get\_current\_stock\_price} &
None &
Return the most recent trading price for \{company\}'s stock, or -1 if unavailable. \\

\addlinespace
\texttt{get\_stock\_price\_history} &
\texttt{timeline} \newline (Literal: \texttt{d}, \texttt{w}, \texttt{m}) &
Retrieve the closing stock prices for \{company\} over the past year with a daily, weekly, or monthly resolution. Returns the last 10 values. \\

\addlinespace
\texttt{get\_analyst\_price\_targets} &
\texttt{target\_type} \newline (Literal: \texttt{current}, \texttt{low}, \texttt{high}, \texttt{mean}, \texttt{median}) &
Fetch a specific analyst price target for \{company\}, such as current, high, low, mean, or median forecasted price. \\

\addlinespace
\texttt{get\_revenue} &
\texttt{year} (Optional: int) &
Get \{company\}'s total revenue by year. If no year is provided, returns all available revenue data. \\

\addlinespace
\texttt{get\_net\_income} &
\texttt{year} (Optional: int) &
Get \{company\}'s net income by year. If no year is specified, returns all available net income data. \\

\end{longtable}
\endgroup

\end{document}